\documentclass[%
superscriptaddress,
preprint,
 amsmath,amssymb,
 aps,
]{revtex4-1}

\usepackage{graphicx}% Include figure files
\usepackage{dcolumn}% Align table columns on decimal point
\usepackage{bm}% bold math
\usepackage{xcolor}
\usepackage{qcircuit}
\usepackage{braket}
\usepackage{float}
\usepackage{algorithm}
\usepackage{algorithmicx}
\usepackage{algpseudocode}
\usepackage{mathtools}
\usepackage[scientific-notation=true]{siunitx}
\usepackage{hyperref}

\usepackage{soul} % Cross through text

\usepackage{listofitems} % for \readlist to create arrays
\usepackage{tikz}
\tikzstyle{mynode}=[thick, draw=blue, fill=blue!20, circle, minimum size=22]

\renewcommand{\figureautorefname}{Figure~\negthinspace}
\renewcommand{\equationautorefname}{Equation~\negthinspace}
\renewcommand{\tableautorefname}{Table~\negthinspace}
\renewcommand{\sectionautorefname}{Section~\negthinspace}

\begin{document}

% \preprint{BNL}

\title{Reservoir Computing via Quantum Recurrent Neural Networks}% Force line breaks with \\

\author{Samuel Yen-Chi Chen}
% \email{ycchen1989@gmail.com}
\affiliation{%
 Wells Fargo
}%
\author{Daniel Fry}
\affiliation{%
 IBM Quantum, IBM Research
}
\author{Amol Deshmukh}
\affiliation{%
 IBM Quantum, IBM Research
}
\author{Vladimir Rastunkov}
\affiliation{%
 IBM Quantum, IBM Research
}
\author{Charlee Stefanski}
\affiliation{%
 Wells Fargo
}

% \collaboration{MUSO Collaboration}%\noaffiliation

%  \homepage{http://www.Second.institution.edu/~Charlie.Author}
% \affiliation{
%  Second institution and/or address\\
%  This line break forced% with \\
% }%
% \affiliation{
%  Third institution, the second for Charlie Author
% \affiliation{%
%  Authors' institution and/or address\\
%  This line break forced with \textbackslash\textbackslash
% }%

% \collaboration{CLEO Collaboration}%\noaffiliation

\date{\today}% It is always \today, today,

\begin{abstract}
Recent developments in quantum computing and machine learning have propelled the interdisciplinary study of quantum machine learning. Sequential modeling is an important task with high scientific and commercial value.
% Challenge
Existing VQC or QNN-based methods require significant computational resources to perform the gradient-based optimization of a larger number of quantum circuit parameters. The major drawback is that such quantum gradient calculation 
% depends on the method called parameter-shift rule that 
requires a large amount of circuit evaluation, posing challenges in current near-term quantum hardware and simulation software.
% Approach
% In this work, we approach sequential modeling by applying a reservoir computing (RC) framework to quantum recurrent neural networks (QRNN-RC). 
In this work, we approach sequential modeling by applying a reservoir computing (RC) framework to quantum recurrent neural networks (QRNN-RC) that are based on classical RNN, LSTM and GRU.
The main idea to this RC approach is that the QRNN with randomly initialized weights is treated as a dynamical system and only the final classical linear layer is trained.
% Results & Impact
Our numerical simulations show that the QRNN-RC can reach results comparable to fully trained QRNN models for several function approximation and time series prediction tasks. Since the QRNN training complexity is significantly reduced, the proposed model trains notably faster. In this work we also compare to corresponding classical RNN-based RC implementations and show that the quantum version learns faster by requiring fewer training epochs in most cases. 
% Our results open a new possibility to utilize quantum neural network for sequential modeling with much fewer resources. 
Our results demonstrate a new possibility to utilize quantum neural network for sequential modeling with greater quantum hardware efficiency, an important design consideration for noisy intermediate-scale quantum (NISQ) computers.
\end{abstract}

\maketitle

%\tableofcontents
% \tableofcontents

\section{\label{sec:Indroduction}Introduction}
%
% \YC{Describe the QC\\}
%
Quantum computing (QC) has been demonstrated theoretically to provide significant speedup over classical computers in several computational tasks \cite{harrow2017quantum, nielsen2002quantum}. Notable examples include the factoring of large numbers \cite{shor1994algorithms} and searching an unstructured database \cite{grover1996fast}. Recent advances in quantum hardware by companies such as IBM \cite{cross2018ibm}, Google \cite{arute2019quantum} and IonQ \cite{debnath2016demonstration} \st{provide} enable the opportunity to implement quantum algorithms on real devices.
% \YC{Mention recent hardware development here\\}
% \YC{Describe the classical ML\\}
%
At the same time, the development of various machine learning (ML) techniques has accelerated progress in fields such as natural language processing \cite{cho2014learning,sutskever2014sequence}, automatic speech recognition \cite{graves2013speech,graves2014towards,sak2015fast,sak2014long}, computer vision \cite{krizhevsky2012imagenet,szegedy2015going,simonyan2014very,lecun1998gradient,he2016deep}, complex sequential decision making \cite{silver2017mastering,silver2016mastering,Mnih2015Human-levelLearning,schrittwieser2019mastering,badia2020agent57} and many more.
% \YC{Describe the QML\\}

%
Considering the ever increasing data volume and complexity of accessible data, it is reasonable to examine whether we can build more powerful ML methods with the help of a novel computing paradigm. QC is a leading candidate and the attempt to address this problem let to the development of quantum machine learning (QML) \cite{dunjko2018machine,biamonte2017quantum}.
%
% \YC{Describe the research field we plan to solve in this paper and mention current methods\\}
Sequential modeling is a common ML task and has been studied extensively in the classical setting. 
% \YC{Cite modern RNN and transformer methods. classical RC methods} 
For example, the recurrent neural network (RNN) \cite{dupond2019thorough,abiodun2018state,tealab2018time} and its variants--such as gated recurrent units (GRU) \cite{cho2014properties} and long short-term memory (LSTM) \cite{hochreiter1997long}--has a long history of being applied in machine translation \cite{cho2014learning,sutskever2014sequence}, speech recognition \cite{graves2013speech,graves2014towards,sak2015fast,sak2014long} and time-series analysis \cite{connor1994recurrent,hua2019deep}, just name a few.
Indeed, sequential modeling has also been studied in the QML field via the use of quantum recurrent networks (QRNN) \cite{bausch2020recurrent,takaki2020learning} and its variants such as quantum long short-term memory (QLSTM) \cite{chen2020quantum}.
%
% \YC{Describe the challenges of existing methods\\}
However, existing methods using QRNN and its variants to study sequential modeling suffers from a major drawback: long training time. QML methods for sequential modeling such as QRNN and QLSTM largely depend on the iterative optimization of quantum circuit parameters. Notable examples are variational quantum algorithms (VQA) \cite{cerezo2021variational} and quantum circuit learning (QCL) \cite{mitarai2018quantum}; both require a significant amount of circuit evaluation to calculate the gradients and update the circuit parameters \cite{schuld2019evaluating}. For example, the commonly used \emph{parameter-shift} quantum gradient calculation method requires two circuit evaluations for each parameter \cite{mitarai2018quantum,schuld2019evaluating}.

Intuitively, one can ask the following question: can we only train part of the model instead of all of the parameters and achieve comparable results? The answer is yes when classical RNNs are randomly initialized to process the sequence and only the final linear layer is trained. Such architecture is called \emph{reservoir computing} (RC) \cite{jaeger2004harnessing,jaeger2001echo,tanaka2019recent}.
%
% \YC{Add a short text here to mention RC can be any blackbox and RNN is one of them}
%
While RC based on classical RNN has demonstrated significant success, as described in \cite{jaeger2004harnessing,jaeger2001echo}, it is not yet clear whether their quantum counterpart (e.g. quantum RNN and variants) can achieve comparable or superior results. In this paper, we propose a reservoir computing (RC) method based on randomly initialized quantum circuits. Specifically, we investigate the quantum version of RNN-based RC. We consider the following quantum RNN: quantum recurrent neural network (QRNN), quantum long short-term memory (QLSTM) and a quantum gated recurrent unit (QGRU). We apply the untrained QRNN, QGRU and QLSTM as the reservoir and only train the final classical linear layer which is used to process the output from the respective quantum reservoirs.
%
% \YC{Describe our results and impacts\\}

The numerical simulations show that the QRNN-RC can reach results comparable to fully trained QRNN models in several function approximation and time-series prediction tasks. Since the QRNNs in the proposed model does not need to be trained, the overall process is much faster than the fully trained ones. We also compare to classical RNN-based RC and show that in most cases the quantum version learns faster or requires fewer training epochs. 
% \YC{Describe the organization of the paper\\}

The paper is organized as follows: In \sectionautorefname{\ref{sec:ReservoirComputing}} the basic notion of reservoir computing is described. In \sectionautorefname{\ref{sec:VQC}} we introduce the VQC which is the building block of QML models. We describe various kinds of QRNNs in the \sectionautorefname{\ref{sec:QRNN}}. The experimental settings are described in \sectionautorefname{\ref{sec:Exp}} and the results are shown in \sectionautorefname{\ref{sec:Results}}. Finally, we discuss the results in \sectionautorefname{\ref{sec:Discussion}} and provide concluding remarks in \sectionautorefname{\ref{sec:Conclusion}}.
%
% \YC{General introduction of the classical machine learning and quantum computing}\\
% Recently, machine learning (or deep learning) has been successful in computer vision, natural language processing and basic science research.\\
% \YC{You will need some literature citation here, you can see my paper or simply Google some paper and list some of the most cited papers in each field.}
% At the meantime, quantum computing, once seen as not realizable, has been brought to the market by several tech giants and startup companies.\\
% \YC{Say something about the quantum computing, again you will need citations.}\\
% However, these machines are not equipped with the feature of quantum error correction and therefore cannot perform \emph{fault-tolerant quantum computation}.\\
% \YC{Elaborate some of the limitations of NISQ, need citations, you may need to say a lot more!}\\
% \YC{Then say about our current research, why is it important? Most of the time it is that no others have been solved or demonstrated this.}\\
% The problem of \textcolor{red}{deep reinforcement learning}, to our best knowledge, has not been investigated in the quantum domain. In this work, we propose a novel framework to demonstrate the feasibility of implementing deep reinforcement learning with \emph{variational quantum circuits} and show that we can harvest certain quantum advantages within this scheme.
%

\section{\label{sec:ReservoirComputing}Reservoir Computing}
%
% \YC{why we need RC}
A fundamental task in machine learning is to model temporal or sequential data. Examples of this include ML models trained to process audio or text data to perform natural language processing \cite{cho2014learning,sutskever2014sequence,graves2013speech,graves2014towards,sak2015fast,sak2014long}, or analyze financial data to provide better decision making \cite{krollner2010financial,dingli2017financial}. Various recurrent neural networks (RNN) are often used to achieve these tasks. However, there are challenges when training RNN such as vanishing or exploding gradients \cite{hochreiter1998vanishing,pascanu2013difficulty}, and training RNNs is usually computationally expensive. 
\begin{figure}[htbp]
\begin{center}
\begin{tikzpicture}[x=2.2cm,y=1.4cm]

\draw[color=black, fill=gray!5, thin](1, 2) circle (1.55 and 2.5);
\foreach \x in {1, 2, 3}
        \draw [black, thick] (-1.5, \x) circle [radius=7pt];
\draw[black, thick] (1, 0) circle [radius=9pt];
\draw[black, thick] (1.5, 1) circle [radius=9pt];
\draw[black, thick] (0.5, 2) circle [radius=9pt];
\draw[black, thick] (1, 3) circle [radius=9pt];
\draw[black, thick] (1.25, 4) circle [radius=9pt];

% final
\draw [black, thick] (3.9, 2.2) circle [radius=9pt];

%arrows
\foreach \x in {1, 3}
    \draw[-stealth, black, thin] (-1.35, \x) -- (1 - 0.2, 0 -0.2 + 0.1 * \x);
\foreach \x in {2}
    \draw[-stealth, black, thin] (-1.35, \x) -- (1.5 - 0.2, 0.8 + 0.1 * \x);
\foreach \x in {2, 3}
    \draw[-stealth, black, thin] (-1.35, \x) -- (1.25 - 0.2, 3.8 + 0.1 * \x);
% \draw[-stealth, black, thick] (-1.3, 2) -- (1 - 0.2, 3 -0.2 + 0.1);
\draw[-stealth, black, thin] (-1.35, 1) -- (0.5 - 0.2, 1.8 + 0.1);

% intermediate layer
\draw[-stealth, black!60!green, thick] (1.5, 1 + 0.25) to [bend right] (1.25, 4 - 0.25);
\draw[-stealth, black!60!green, thick] (1.5 - 0.1, 1 + 0.25) to [bend left] (0.5 + 0.2, 2 - 0.25);
\draw[-stealth, black!60!green, thick] (1, 0.3) to [bend right=10] (1, 2.7);
\draw[-stealth, black!60!green, thick] (0.6, 2.2) to [bend left] (0.9, 2.8);
\draw[-stealth, black!60!green, thick] (1.65, 0.8) to [bend right=-40, in=35, out=120, looseness=4] (1.35, 0.8);
\draw[-stealth, black!60!green, thick] (0.85, 3.2) to [bend left=-40, in=35, out=100, looseness=4] (1.1, 3.2);

% final layer
\draw[-stealth, blue, dashed, thick] (1 + 0.15, 0) to [bend right=10] (3.9 - 0.2, 2 + 0.1 * 0);
\draw[-stealth, blue, dashed, thick] (1 + 0.15, 3) -- (3.9 - 0.2, 2 + 0.1 * 2);
\draw[-stealth, blue, dashed, thick] (1.25 + 0.15, 4) to [bend right=-10] (3.9 - 0.2, 2 + 0.1 * 4);

% symbols
\draw(-.9, 3.4) node{${W^{in}}$};
\draw(2.9, 3.5) node{${W^{out}}$};
\draw(1.8, 1.95) node{${W}$};

\draw(-1.9, 2) node{${s_k}$};
\draw(1, -0.75) node{${\textbf{x}_k}$};
\draw(4.3, 2.2) node{${y_k}$};
\end{tikzpicture}
\end{center}
\caption{{\bfseries Reservoir computing (RC).}}
\label{Fig:classical_reservoir_computing}
\end{figure}
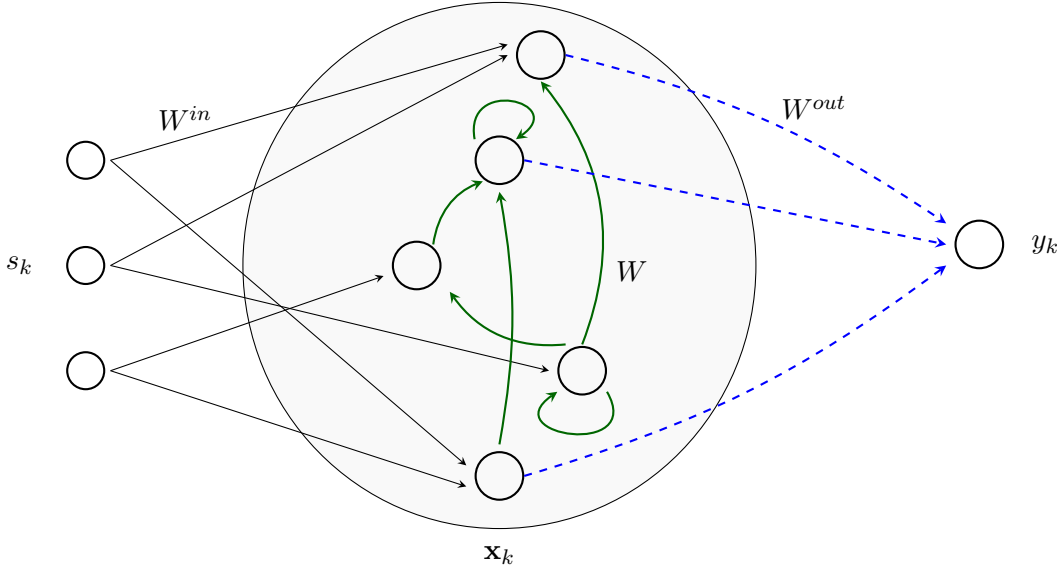

%
% \YC{vlad}
Reservoir computing (RC) is defined in \cite{Miikkulainen2017} as an approach to processing sequential data, where large, nonlinear, randomly connected, and fixed recurrent network (the \emph{reservoir}) is separated from a linear output layer with trainable parameters. It is assumed that the complexity of the recurrent network allows one to learn the desired output by using only a linear combination of its activations \cite{jaeger2004harnessing}. The linear output layer is fast to train, so it helps to mitigate issues with RNN training discussed above. RC based on RNN, as depicted in \figureautorefname{\ref{Fig:classical_reservoir_computing}, is sometimes referred to as the echo state network \cite{jaeger2001echo}. It can be summarized mathematically as follows:\begin{align}\mathbf{x}_k &= \mathbf{f}(W^{in} s_k + W \mathbf{x}_{k-1})\nonumber\\y_k &= W^{out} \textbf{x}_k^{out},\end{align}where $s_k$and  $x_k$  correspond to the input signal, and state of the reservoir, respectively, at step $k$. Here, $W$, $W^{in}$, and $W^{out}$ correspond to the internal weights of the reservoir, the weights connecting the input nodes to the nodes in the reservoir, and the weights connecting the reservoir nodes to the output nodes, respectively. Only $W^{out}$ needs to be trained, other weights are randomly initialized.
With the success of classical RNN-based reservoirs, it is natural to consider a similar idea in the quantum regime. Specifically, we consider the quantum version of common RNN architectures such as quantum RNN, quantum long short-term memory (QLSTM) and quantum gated recurrent unit (QGRU). 
Along with the idea of classical RNN-based RC, we replaced the classical neural networks inside these RNN architectures with variational quantum circuits (VQC) which have been shown to have certain advantages over classical neural networks \cite{caro2022generalization,du2018expressive,abbas2021power}.
In the next section, we will describe the building blocks of these quantum RNNs.
\section{\label{sec:VQC}Variational Quantum Circuits}
A variational quantum circuit (VQC) (also known as a parameterized quantum circuit (PQC)), is a quantum circuit which depends on tunable parameters. The parameters can be tuned via gradient-based \cite{schuld2019evaluating,pellow2021comparison} or gradient-free algorithms \cite{franken2020gradient,pellow2021comparison}. \figureautorefname{\ref{Fig:GeneralVQC}} illustrates a generic VQC which consists of three parts: state preparation, the parameterized circuit, followed by measurement. In the figure, $U(\mathbf{x})$ represents the state preparation circuit which encodes classical data $\mathbf{x}$ into a quantum state. $V(\boldsymbol{\theta})$ represents the variational or parameterized circuit with \emph{learnable} or adjustable parameters $\boldsymbol{\theta}$, which, in the context of this paper, is optimized using gradient-descent. The output is obtained as a classical bit string through measurement of a subset (or all) of the qubits.

\begin{figure}[hbtp]
\begin{center}
\scalebox{1.4}{
\begin{minipage}{10cm}
\Qcircuit @C=1em @R=1em {
\lstick{\ket{0}} & \multigate{3}{U(\mathbf{x})}  & \qw        & \multigate{3}{V(\boldsymbol{\theta})}       & \qw      & \meter \qw \\
\lstick{\ket{0}} & \ghost{U(\mathbf{x})}         & \qw        & \ghost{V(\boldsymbol{\theta})}              & \qw      & \meter \qw \\
\lstick{\ket{0}} & \ghost{U(\mathbf{x})}         & \qw        & \ghost{V(\boldsymbol{\theta})}              & \qw      & \meter \qw \\
\lstick{\ket{0}} & \ghost{U(\mathbf{x})}         & \qw        & \ghost{V(\boldsymbol{\theta})}              & \qw      & \meter \qw \\
}
\end{minipage}
}
\end{center}
% \caption[Generic circuit architecture for the variational quantum circuit.]{{\bfseries Generic circuit architecture for the variational quantum circuit.}
\caption{{\bfseries Generic architecture for variational quantum circuits (VQC).}
%  deep reinforcement learning.} This is a generic variational quantum circuit architecture for deep $Q$ network (VQ-DQN).
$U(\mathbf{x})$ is a quantum circuit for encoding the classical input data $\mathbf{x}$ into a quantum state and $V(\boldsymbol{\theta})$ is the variational circuit with tunable or learnable parameters $\boldsymbol{\theta}$ which is optimized via gradient-based or gradient-free methods. This circuit is followed by measurement of some or all of the qubits.
}
\label{Fig:GeneralVQC}
\end{figure}
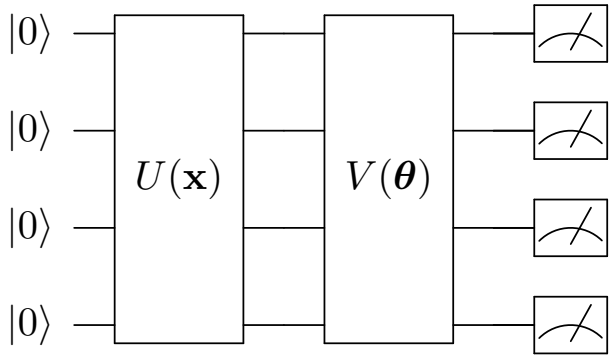

Noteworthy advantages of VQCs include resilience of quantum noise \cite{kandala2017hardware,farhi2014quantum,mcclean2016theory}, which makes them favorable for NISQ era quantum devices, and the ability to train VQCs with smaller datasets \cite{caro2022generalization}. 
% \YC{Need to update this list based on google scholar records\\}
Quantum machine learning methods using VQCs demonstrate a varying degree of success. Notable examples VQC applications include function approximation \cite{chen2020quantum, mitarai2018quantum}, classification \cite{mitarai2018quantum,schuld2018circuit,havlivcek2019supervised,Farhi2018ClassificationProcessors,benedetti2019parameterized,mari2019transfer, abohashima2020classification, easom2020towards, sarma2019quantum, stein2020hybrid,chen2020hybrid,chen2020qcnn,wu2020application,stein2021quclassi,chen2021hybrid,jaderberg2021quantum,mattern2021variational,qi2021qtn,kyriienko2022unsupervised,li2022quantum,wu2022scalable,nguyen2022bayesian}, generative modeling \cite{dallaire2018quantum,stein2020qugan, zoufal2019quantum, situ2018quantum,nakaji2020quantum}, deep reinforcement learning \cite{chen19, chen2022variational, lockwood2020reinforcement,jerbi2019quantum,Chih-ChiehCHEN2020,wu2020quantum,skolik2021quantum,jerbi2021variational,hsiao2022unentangled,yun2022quantum,sequeira2022variational,heimann2022quantum,schenk2022hybrid,chen2022quantum}, sequence modeling \cite{chen2020quantum,bausch2020recurrent, takaki2020learning}, speech recognition \cite{yang2020decentralizing,qi2022classical}, natural language processing \cite{yang2022bert,di2022dawn}, metric and embedding learning \cite{lloyd2020quantum, nghiem2020unified}, transfer learning \cite{mari2019transfer} and federated learning \cite{chen2021federated,yang2020decentralizing,chehimi2021quantum}.
Additionally, it has been shown that the VQCs may have more expressive power than classical neural networks \cite{sim2019expressibility,lanting2014entanglement,du2018expressive,abbas2021power}. The \emph{expressive power} is defined by the ability to represent certain functions or distributions given a limited number of parameters or a specified model size. 
Indeed, artificial neural networks (ANNs) are known as \emph{universal approximators} \cite{hornik1989multilayer}, i.e. a neural network with even one single hidden layer can, in principle, approximate any computable function. However, as the complexity of the function grows, the number of neurons required in the hidden layer(s) may become extremely large, increasing the demand for computational resources. Thus, it is worthwhile to examine whether VQCs can perform better than their classical counterparts with an equally limited number of parameters.

%%%%%%%
In the optimization procedure, we employ the \emph{parameter-shift} method to derive the analytical gradient of the quantum circuits, as described in \cite{schuld2019evaluating,bergholm2018pennylane}.
%
% \YC{We need to mention that we consider two different approaches here: first one is the traditional full optimization and the other is the reservoir computing. }
In this paper, VQCs are operated in the following ways: (i) in the reservoir computing cases, the VQCs are randomly initialized and then the parameters are fixed, no quantum gradients are needed in this case. (ii) In the full optimization cases, the VQCs are optimized through gradient-based methods.
In the next section, we describe the quantum version of RNNs used in this work.
\section{\label{sec:QRNN}Quantum Recurrent Neural Network}
RNNs are a special kind of ML model designed to handle sequential modeling via the memory capabilities which can keep track of previous information. What makes RNNs and its variants special is that the output from the RNN will be fed into the model again to retain previous information. The value fed back to the RNN is called the \emph{hidden} state. This is the major difference between a RNN and a fully-connected neural network.
%
% \YC{Give an example of RNN operation: both sequence modeling or prediction.}
%
RNNs can be used to learn and output a whole sequence or predict a single value. In the first case, at each time step $t$, given the hidden state from the previous time $h_{t-1}$ and the input $x_{t}$, the RNN will output the prediction $y_{t}$ and the hidden state $h_{t}$. In the other case, if we choose to use the RNN to predict a single value, then given an input sequence $\{x_0, x_1, \cdots, x_n\}$, only the final $y_{n}$ will be retained.
%

%
% \YC{FIG: RNN diagram-sequence learning\\}
% \YC{FIG: RNN diagram-single value prediction\\}
The generic form of a RNN suffers from several challenges such as vanishing gradients \cite{hochreiter1998vanishing,pascanu2013difficulty} and failing to learn long-range temporal dependencies \cite{hochreiter1998vanishing,pascanu2013difficulty}. Various modified forms of RNNs have been proposed to fix these issues such as long short-term memory (LSTM) \cite{hochreiter1997long} and gated recurrent units (GRU) \cite{cho2014properties}, which have demonstrated superior performance over the generic RNN in a wide range of applications \cite{salehinejad2017recent}.
RNN and its variants such as LSTM and GRU can be used to serve as a high-dimensional dynamical system or as a \emph{reservoir}. In this case, the RNN is not trained, meaning that its parameters are fixed after the random initialization \cite{lukovsevivcius2009reservoir}. The only trainable part is the final linear layer which will process the output from the RNN.
%
% \YC{FIG: RNN-RC diagram\\}
%

%
\subsection{Quantum Recurrent Neural Network}
\begin{figure}[htbp]
\includegraphics[width=0.6\linewidth]{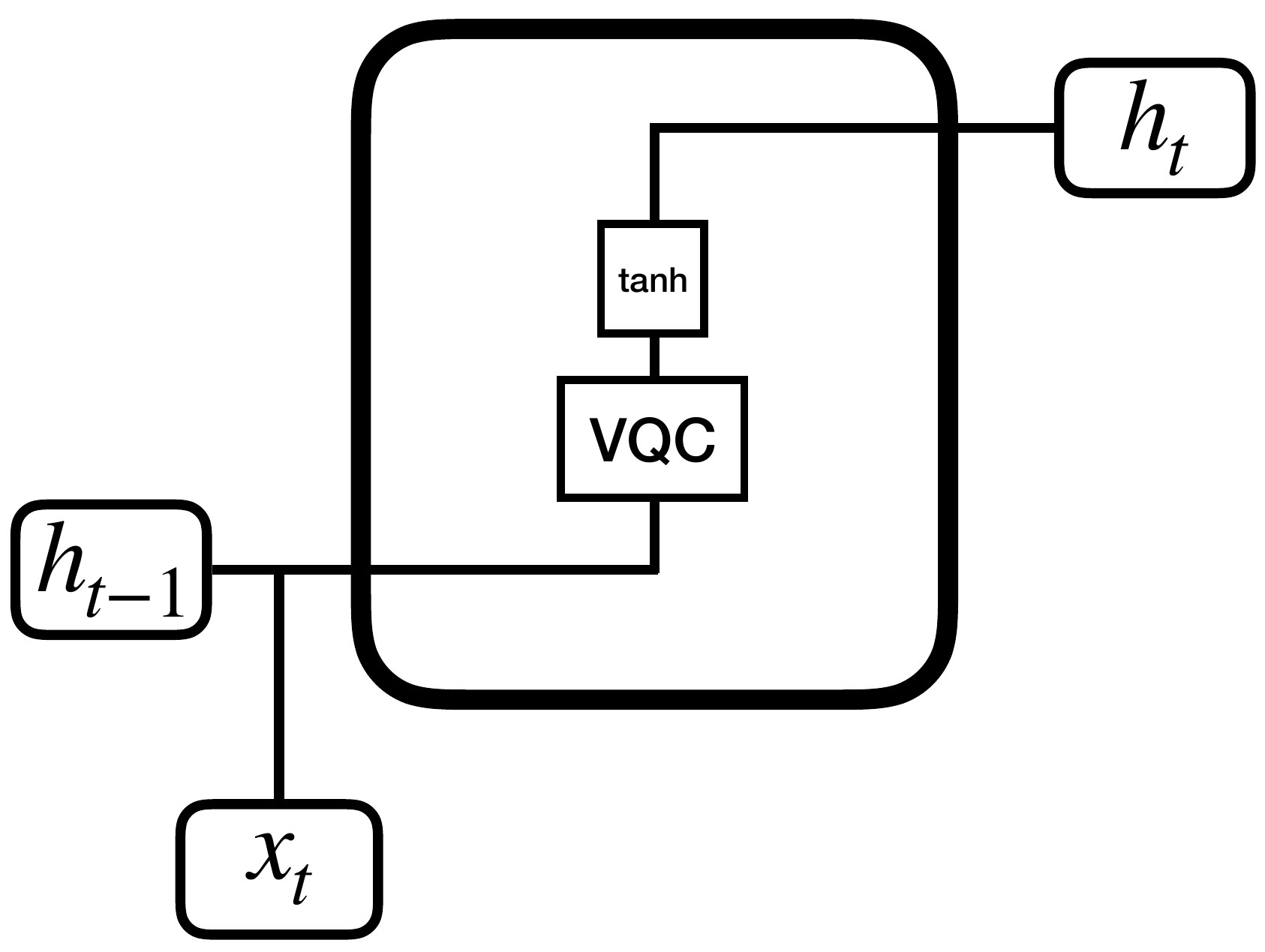}% Here is how to import EPS art
\caption{{\bfseries The quantum recurrent neural networks (QRNN) architecture.} }
\label{fig:QRNN}
\end{figure}
The quantum recurrent neural network (QRNN) is the quantum version of the conventional RNN. The major distinction is that the classical neural network is replaced by a VQC, as shown in \figureautorefname{\ref{fig:QRNN}}. The formulation of a QRNN cell is given by

\begin{subequations}
\allowdisplaybreaks
    \begin{align}
    h_{t} &= \tanh(VQC(v_{t})) 
    \label{eqn:qrnn-h}\\
    y_{t} &= NN(h_{t}) \label{eqn:qrnn-yt}
    \end{align}
    \label{eqn:qrnn}
\end{subequations}
where the input is the concatenation $v_t$ of the hidden state $h_{t-1}$ from the previous time step and the current input vector $x_t$. The VQC is detailed in the \sectionautorefname{\ref{sec:VQCcomponentsForQRNN}}. In this work, $x_t$ is set to be one-dimensional and the hidden unit $h_{t}$ is set to be three-dimensional.
Since the model is built to generate the prediction of a scalar value, the output from the QRNN, $h_{t}$, at the last time step (in the context of this paper the last step is $t = 4$) will be processed by a classical neural network layer $NN$ (as in \equationautorefname{\ref{eqn:qrnn-yt}}).
%
% \YC{QRNN diagram\\}
% \YC{Check the equation and the code again\\}
%

%
\subsection{Quantum Long Short-term Memory}
\begin{figure}[htbp]
\includegraphics[width=0.6\linewidth]{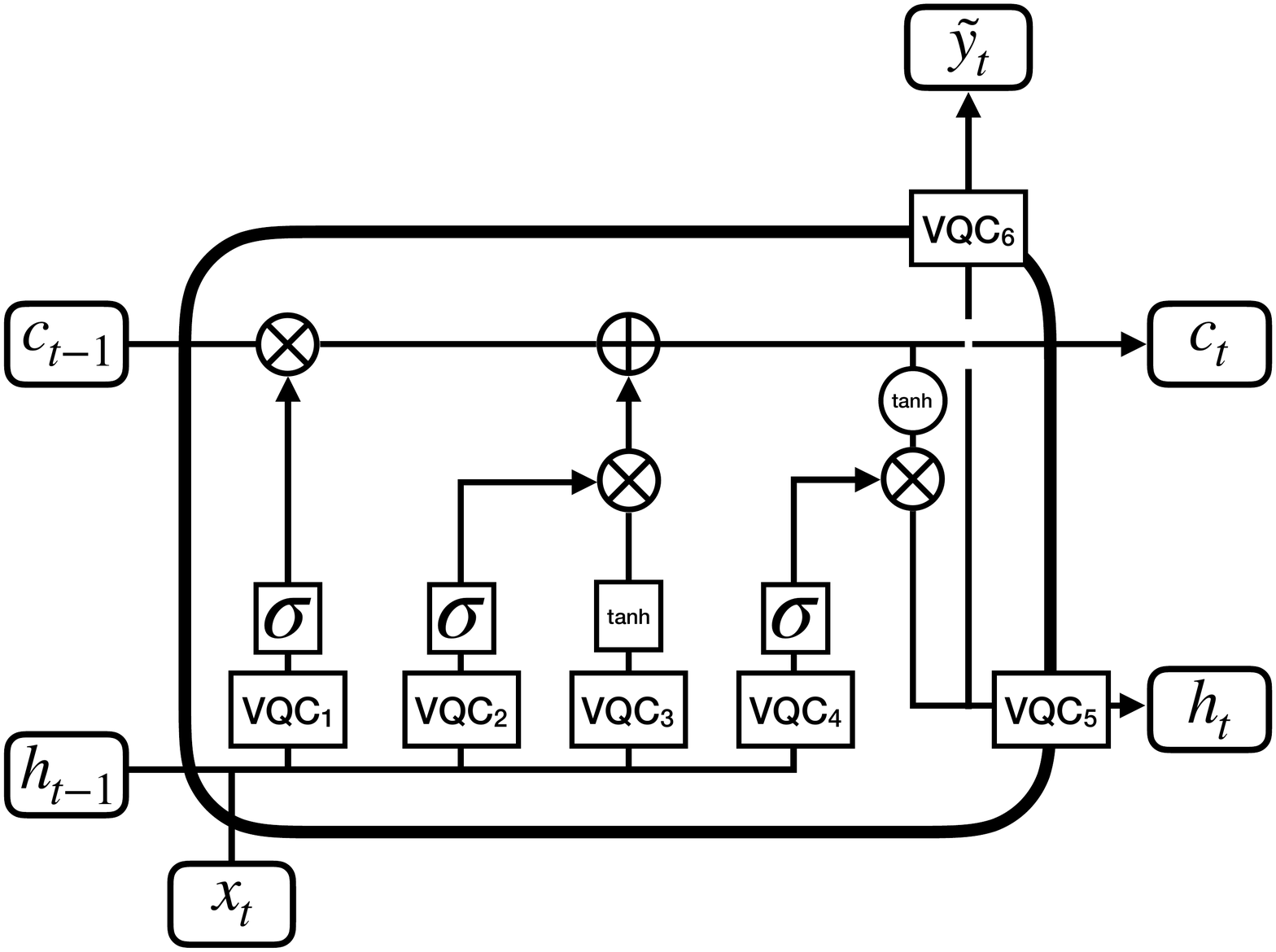}% Here is how to import EPS art
\caption{{\bfseries The quantum long short-term memory (QLSTM) architecture.} }
\label{fig:QLSTM}
\end{figure}
The quantum long short-term memory (QLSTM) \cite{chen2020quantum} is an improved version of QRNN. There are two memory components in a QLSTM, namely the hidden state $h_t$ and the cell or internal state $c_t$.
A formal mathematical formulation of a QLSTM cell is given by
\begin{subequations}
\allowdisplaybreaks
    \begin{align}
    f_{t} &= \sigma\left(VQC_{1}(v_t)\right) \label{eqn:qlstm-f}\\
    i_{t} &= \sigma\left(VQC_{2}(v_t)\right) \label{eqn:qlstm-i}\\ 
    \tilde{C}_{t} &= \tanh \left(VQC_{3}(v_t)\right) \label{eqn:qlstm-bigC}\\
    c_{t} &= f_{t} * c_{t-1} + i_{t} * \tilde{C}_{t} \label{eqn:qlstm-c}\\
    o_{t} &= \sigma\left(VQC_{4}(v_t)\right) \label{eqn:qlstm-o}\\ 
    h_{t} &= VQC_{5}(o_{t} * \tanh \left(c_{t}\right)) \label{eqn:qlstm-h}\\
    \tilde{y_{t}} &= VQC_{6}(o_{t} * \tanh \left(c_{t}\right)), \label{eqn:qlstm-y}\\
    y_{t} &= NN(\tilde{y_{t}}) \label{eqn:qlstm-final-y}
    \end{align}
    \label{eqn:qlstm}
\end{subequations}
where the input is the concatenation $v_t$ of the hidden state $h_{t-1}$ from the previous time step and the current input vector $x_t$. The VQC is detailed in the \sectionautorefname{\ref{sec:VQCcomponentsForQRNN}}. In this work, the $x_t$ is set to be one dimensional and the hidden unit $h_{t}$ is set to be three dimensional. The cell state or internal state $c_{t}$ is set to be four-dimensional.
Since the model is built to generate the prediction of a scalar value, the output from the QLSTM $\tilde{y_{t}}$ at the last time step  (in the context of this paper the last step is $t = 4$) will be processed by a classical neural network layer $NN$ to get $y_{t}$.
%
% \YC{QLSTM diagram\\}
% \YC{Check the equation and the code again\\}
%
\subsection{Quantum Gated Recurrent Unit}
\begin{figure}[htbp]
\includegraphics[width=0.6\linewidth]{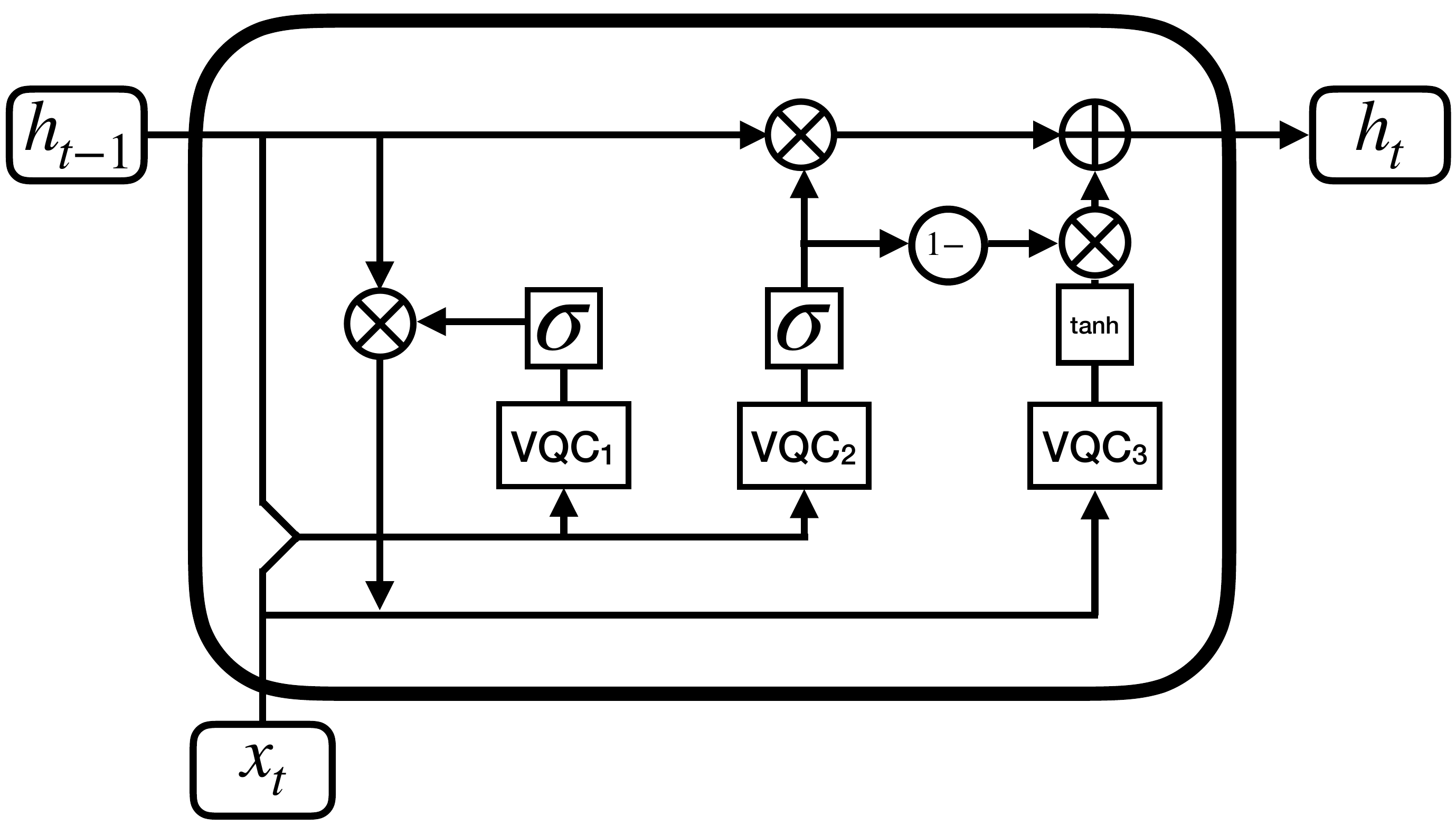}% Here is how to import EPS art
\caption{{\bfseries The quantum gated recurrent units (QGRU) architecture.} }
\label{fig:QGRU}
\end{figure}
The quantum gated recurrent unit (QGRU) is another QRNN with gating mechanisms similar to QLSTM. QGRU has fewer parameters and simpler architectures than QLSTM. 
A formal mathematical formulation of a QGRU cell is given by
\begin{subequations}
\allowdisplaybreaks
    \begin{align}
    r_{t} &= \sigma\left(VQC_{1}(v_t)\right) \label{eqn:qgru-rt}\\
    z_{t} &= \sigma\left(VQC_{2}(v_t)\right) \label{eqn:qgru-zt}\\
    o_{t} &= cat(x_{t}, r_{t} * H_{t-1}) \label{eqn:qgru-ot}\\
    \tilde{H}_{t} &= \tanh \left( VQC_{3}(o_{t}) \right) \label{eqn:qgru-tilda-Ht}\\
    H_{t} &= z_{t} * H_{t-1} + (1 - z_{t}) * \tilde{H}_{t} \label{eqn:qgru-Ht}\\
    y_{t} &= NN(H_{t}) \label{eqn:qgru-yt}
    \end{align}
    \label{eqn:qgru}
\end{subequations}
where the input is the concatenation $v_t$ of the hidden state $H_{t-1}$ from the previous time step and the current input vector $x_t$. The VQC is detailed in the \sectionautorefname{\ref{sec:VQCcomponentsForQRNN}}. In this work, the $x_t$ is set to be one-dimensional and the hidden unit $H_{t}$ is set to be three-dimensional.
Since the model is built to generate the prediction of a scalar value, the output from the QGRU $H_{t}$ at the last time step  (in the context of this paper the last step is $t = 4$) will be processed by a classical neural network layer $NN$ to get $y_{t}$.
%
% \YC{QGRU diagram\\}
% \YC{Check the equation and the code again\\}
% \YC{Do we need to put NN in the formula? Or mention in the main text}
%
\subsection{VQC Components}
\label{sec:VQCcomponentsForQRNN}
% \YC{This part describe the VQC components: encoding, variational and measurement.}
%
The specific VQC components used in this paper are represented in \figureautorefname{\ref{Fig:Basic_VQC_Hadamard_MoreEntangle}}. As previously mentioned, a VQC includes the following three parts: an \emph{encoding circuit}, a \emph{variational circuit} and \emph{quantum measurement}.
\subsubsection{Encoding Circuit}
A quantum state with $N$-qubits can be defined as
\begin{equation}
\label{eqn:quantum_state_vec}
    \ket{\psi} = \sum_{(q_1,q_2,\cdots,q_N) \in \{ 0,1\}} c_{q_1, q_2, \cdots, q_N}\ket{q_1} \otimes \ket{q_2} \otimes \cdots \otimes \ket{q_N},
\end{equation}
where $ c_{q_1, \cdots, q_N} \in \mathbb{C}$ is the complex \emph{amplitude} for each basis state and $q_i \in \{0,1\}$. 
The square of the amplitude $c_{q_1, \cdots, q_N}$ is the measurement \emph{probability} for the corresponding value in  $\ket{q_1} \otimes \ket{q_2} \otimes \cdots \otimes \ket{q_N}$, such that the total probability is $1$:

\begin{equation} 
\label{eqn:quantum_state_vec_normalization_condition}
\sum_{(q_1, \cdots, q_N) \in \{0, 1\}} ||c_{q_1, \cdots, q_N}||^2 = 1. 
\end{equation}
%
% The \emph{encoding} circuit is designed to transform the classical inputs into those quantum amplitudes. In this paper, we follow the encoding procedure used in the work \cite{chen2020quantum}. Consider a four dimensional input $x_{1}, \cdots x_{4}$ to the circuit. The circuit is initialized with all zeros and then the Hadamard gate is applied to create unbiased initial state. We then apply $R_{y}$ and $R_{z}$ on $i$-th qubit with the angles $\arctan(x_{i})$ and $\arctan(x_{i}^{2})$ respectively.
The encoding circuit maps classical data values to quantum amplitudes. In this paper, we use the encoding procedure described in \cite{chen2020quantum}. The circuit is initialized in the ground state and then Hadamard gates are applied to create an unbiased initial state. We use a two-angle encoding, similar to dense angle encoding \cite{larose2020robust}, but for encoding one value with two angles. This involves encoding each data value to a qubit with a series of two gates, $R_{y}$ and $R_{z}$, respectively. The angles of the rotation gates are given by $f(x_i) = \arctan(x_i)$ and $g(x_i) = \arctan(x_{i}^{2})$, respectively, where $x_i$ is a component of data vector $\bold{x}$. The quantum state of the encoded data takes the form

\begin{equation}
	\label{eqn:dense_angle_encoding}
	\ket{\bold{x}} = \bigotimes_{i=1}^{N}\,\cos\left(f(x_{i})+\frac{\pi}{4}\right)\ket{0} + \exp{(ig(x_{i}))} \sin\left(f( x_{i})+\frac{\pi}{4}\right)\ket{1}
\end{equation}

where $N$ is the dimensionality of $\bold{x}$ and the $\pi/4$ angle offset accounts for the initial Hadamard rotations.

\subsubsection{Variational Circuit}
The trainable (or learnable) part of the VQC is the \emph{variational} circuit. This is a parameterized circuit where the parameters are subject to iterative optimization, such as gradient-descent. In this paper, the variational part includes several \emph{blocks}, represented as dashed boxes in \figureautorefname{\ref{Fig:Basic_VQC_Hadamard_MoreEntangle}. Each block consists of multiple CNOT gates to entangle qubits, and unitary rotation gates controlled by learnable parameters $\alpha$, $\beta$ and $\gamma$. The blocks can be repeated several times to increase the number of parameters.
\subsubsection{Quantum Measurement}
Our hybrid quantum-classical architecture relies on the ability to move data between quantum and classical systems. To extract the information from the quantum circuit, we perform quantum measurements. Consider the circuit shown in \figureautorefname{\ref{Fig:Basic_VQC_Hadamard_MoreEntangle}} as an example, if we run the circuit once, we will get a bit string like 0011 since we measure all the four qubits. Due to the probabilistic nature of quantum systems, we will get different bit strings at each circuit repetition and measurement. In the next run, it may be, for example, 0110. If we run the circuit many times (number of \emph{shots}), we can get a distribution of the measurement results, called the \emph{expectation values} of the observable. The expectation values can be calculated analytically when using a quantum simulator software without noise, or multiple sampling when a certain device noise model is specified.
%
% \DES{Additional lines}
Given an operator $\hat{O}$, the expected value for a state $|\psi\rangle$ is given by 

\begin{equation}
    \mathbb{E}[\hat{O}] = \langle\psi|\hat{O}|\psi\rangle. 
\end{equation}

In our case, $\ket{\psi}$ corresponds to the state $U\ket{\bold{x}}$ in which $\ket{\bold{x}}$ is the encoded data vector as defined in Equation~\ref{eqn:dense_angle_encoding}, and $U$ is the variational circuit. 
% \DES{Addition ends}
%
% \YC{Continue here.....}
%
%
% \YC{describe the measurement}
% \YC{describe the shots}
% \YC{describe the difference between single circuit running and expectation values.}
%
% \YC{details of the VQC.\\}
%
% \YC{FIG: The VQC used in the QRNNs.}
%
\begin{figure}[htbp]
\begin{center}
\begin{minipage}{10cm}
\Qcircuit @C=1em @R=1em {
\lstick{\ket{0}} & \gate{H} & \gate{R_y(\arctan(x_1))} & \gate{R_z(\arctan(x_1^2))} & \ctrl{1}   & \qw       & \qw      & \targ    & \ctrl{2}   & \qw      & \targ    & \qw      & \gate{R(\alpha_1, \beta_1, \gamma_1)} & \meter \qw \\
\lstick{\ket{0}} & \gate{H} & \gate{R_y(\arctan(x_2))} & \gate{R_z(\arctan(x_2^2))} & \targ      & \ctrl{1}  & \qw      & \qw      & \qw        & \ctrl{2} & \qw      & \targ    & \gate{R(\alpha_2, \beta_2, \gamma_2)} & \meter \qw \\
\lstick{\ket{0}} & \gate{H} & \gate{R_y(\arctan(x_3))} & \gate{R_z(\arctan(x_3^2))} & \qw        & \targ     & \ctrl{1} & \qw      & \targ      & \qw      & \ctrl{-2}& \qw      & \gate{R(\alpha_3, \beta_3, \gamma_3)} & \meter \qw \\
\lstick{\ket{0}} & \gate{H} & \gate{R_y(\arctan(x_4))} & \gate{R_z(\arctan(x_4^2))} & \qw        & \qw       & \targ    & \ctrl{-3}& \qw        & \targ    & \qw      & \ctrl{-2}& \gate{R(\alpha_4, \beta_4, \gamma_4)} & \meter \gategroup{1}{5}{4}{13}{.7em}{--}\qw 
}
\end{minipage}
\end{center}
\caption{{\bfseries Generic VQC architecture for QRNN, QLSTM and QGRU.}
%  deep reinforcement learning.} This is a generic variational quantum circuit architecture for deep $Q$ network (VQ-DQN).
%   The single-qubit gates $R_y(\arctan(x_i))$ and $R_z(\arctan(x_i^2))$ represent rotations along $y$-axis and $z$-axis by the given angle $\arctan(x_i)$ and $\arctan(x_i^2)$, respectively. The choose of arc tangent function is that in general the input values are not in the interval of $[-1, 1]$. The CNOT gates are used to entangle quantum states from each qubit and $R(\alpha,\beta,\gamma)$ represents the general single qubit unitary gate with three parameters.
%   The parameters labeled $R_y(\arctan(x_i))$ and $R_y(\arctan(x_i^2))$ are for state preparation and are not subject to iterative optimization.  Parameters labeled $\alpha_i$, $\beta_i$ and $\gamma_i$ are the ones for iterative optimization.
The VQC we use for QRNN, QLSTM and QGRU includes the following three parts: the data encoding circuit (with $H$, $R_y$, and $R_z$ gates), the variational or parameterized circuit (shown within the dashed outline), and the measurement. Note that the number of qubits and the number of measurements can be adjusted to fit the problem of interest (various input and output dimensions), and the variational layer can contain several iterations to increase the model size or the number of parameters, depending on the capacity and capability of the quantum machine or quantum simulation software used for the (actual or numerical) experiments. In the context of this paper, the number of qubits used is $4$.}
\label{Fig:Basic_VQC_Hadamard_MoreEntangle}
\end{figure}
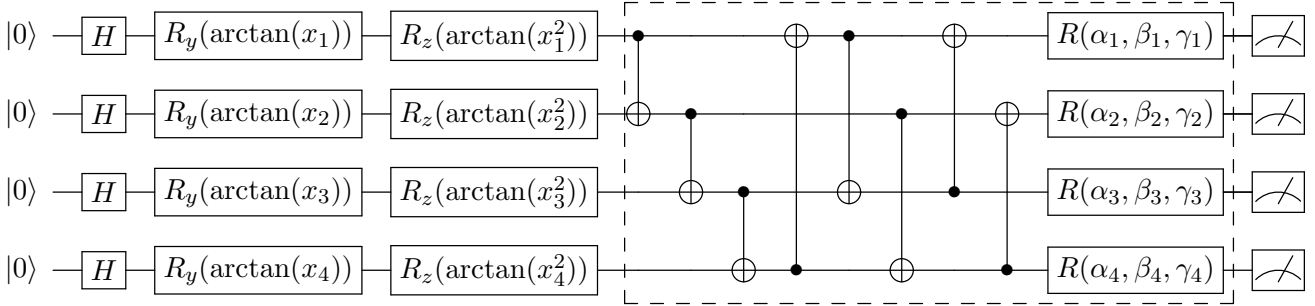
%
%
% \section{Machine Learning}
% \YC{Basic introduction some concepts of machine learning related to this paper.}\\
% \YC{For example, you do a variational quantum reinforcement learning, then you will need to introduce the classical reinforcement learning.}\\
% \YC{For example, you do a variational quantum ODE learning, then you will need to introduce the classical ODE learning.}
% \YC{For example, you do a variational quantum XXX learning, then you will need to introduce the classical XXX learning.}

% \section{Quantum Machine Learning}
% \YC{In this paragraph, we start the exploration of quantum machine learning.}\\
% \YC{Introduce the circuit architecture. Need a figure to describe the generic circuit architecture.} \\
% \YC{In the figure, you need to specify the number of parameters.} \\
% \YC{In the figure, you need to specify the state preparation part and the variational part.} \\
% \YC{Introduce the data-encoding method. Basically there are two encoding methods, amplitude encoding and variational encoding.}\\
% \YC{In the data-encoding part, give some examples for understanding.}\\
% \YC{Introduce the optimization procedure.}\\
% \YC{basically we use the parameter-shift methods. so it is important to cite the related reference.}
% In the optimization procedure, we employ the \emph{parameter-shift} method to derive the analytical gradient of the quantum circuits. The method is described in the reference~\cite{schuld2019evaluating,bergholm2018pennylane}.
%
\section{\label{sec:Exp}Numerical Experiments}
We compare the performance of full optimization and reservoir computing as well as the effect of quantum device noise. In the optimization procedure of quantum circuits, we employ the \emph{parameter-shift} method to derive the analytical gradient of quantum parameters. The method is described in \cite{schuld2019evaluating,bergholm2018pennylane}.
Additionally, we compare our quantum models to classical models with a similar number of parameters. We present the learning performance of these models at different numbers of training epochs.
%
% \YC{Need to mention we carried out the similar experiments as in QLSTM for comparison and cite the paper.}
For a better comparison with previous works, the experimental setting follows that in\cite{chen2020quantum}. We reproduce similar results of fully trained QLSTM and apply the same procedure to QRNN and QGRU.
We use PyTorch \cite{paszke2019pytorch} for the overall ML workflow, PennyLane \cite{bergholm2018pennylane} for building the quantum circuits and Qiskit \cite{cross2018ibm} for noisy quantum simulation.
%
% \YC{Mention the sliding window and dataset split here.\\}
%
% \YC{continue on this part...}

The training and testing scheme follows that in \cite{chen2020quantum}. Concisely, the model is expected to predict the ($N+1$)-th value given the first $N$ values in the sequence. For function approximation tasks (described in \sectionautorefname{\ref{sec:function_approximation_damped_SHM}} and \sectionautorefname{\ref{sec:function_approximation_bessel}), at step $t$ if the input is $[x_{t-4}, x_{t-3}, x_{t-2}, x_{t-1}]$ (i.e., $N = 4$), then the model is expected to generate the output $y_t$, which should be close to the ground truth $x_{t}$. 
For time series prediction tasks (described in \sectionautorefname{\ref{sec:time_series_prediction}}), at step $t$, if the input is $[u_{t-4}, u_{t-3}, u_{t-2}, u_{t-1}]$ (i.e., $N = 4$) from the \emph{input sequence}, then the model is expected to generate the output $y_t$, which should be close to the ground truth $v_{t}$ in the \emph{target sequence}. We set $N=4$ for all experiments in this paper.
\subsection{\label{subsec:full_opt_methods}Full Optimization}
In this part, we present the full optimization (e.g. training all the quantum parameters) of the QRNN to be referenced as the baseline. We consider the QRNN, QGRU and QLSTM models with the following model configurations: For function approximation tasks such as damped SHM and Bessel function, QRNN is with $1 \times 2 \times 4 \times 3 = 24$ trainable quantum parameters and $3 \times 1 + 1 = 4$ trainable classical parameters; QGRU is with $3 \times 2 \times 4 \times 3 = 72$ trainable quantum parameters and $3 \times 1 + 1 = 4$ trainable classical parameters; QLSTM is with $6 \times 2 \times 4 \times 3 = 144$ trainable quantum parameters and $4 \times 1 + 1 = 5$ trainable classical parameters. For time-series prediction tasks such as NARMA5 and NARMA10, the number of trainable quantum parameters are $48$, $144$ and $288$ for QRNN, QGRU and QLSTM, respectively.
%
% These hybrid quantum models are constructed with the following open-source packages: for the quantum part, we adopt the PennyLane \cite{bergholm2018pennylane} and for the classical part, we use the PyTorch \cite{paszke2019pytorch}. 

The optimizer used for this experiment is RMSprop \cite{Tieleman2012}, a variant of gradient descent methods with an adaptive learning rate. The optimizer is configured with the following hyperparameters: learning rate $\eta =0.01$, smoothing constant $\alpha = 0.99$, and $\epsilon = 10^{-8}$.
\subsection{Reservoir Computing}
The RC experiments are configured with the same hyparparameters as the full optimization cases in \sectionautorefname{\ref{subsec:full_opt_methods}}. The only difference is that all the quantum parameters are frozen after the random initialization. Therefore, only classical parameters are trainable.
\subsection{Noisy Simulation}
\label{sec:noise_simulation_definition}
%
% \YC{For IBM collaborators.}
% \textcolor{red}{\textit{\textsc{\underline{DF}: text start}}}
% \\
We used a noise model consisting of serial thermal relaxation and depolarization noise channels, an approach supported by \cite{georgopoulos2021modeling} \cite{dahlhauser2021modeling}. We use high-performance noise model parameters that are largely based on the upper limit performance of the IBM Peekskill superconducting quantum device, currently in exploratory mode. For the thermal relaxation noise channel, T1 and T2 coherence times were sampled per qubit from $\mathcal{N}(500 \mu s, 50 \mu s)$ and $\mathcal{N}(400 \mu s, 40 \mu s)$, respectively, where $\mathcal{N}(\mu,\sigma)$ denotes a normal distribution. Quantum gate and instruction times were fixed values of 0 ns for the $R_Z$ virtual gate, 20 ns for X90 gate, 300 ns for CNOT gate, 700 ns for measurement and 800 ns for reset instruction. Depolarization channel parameters, single-qubit errors and CNOT gate errors were sampled from
$\mathcal{N}(\num{1e-4},\num{1e-5})$ and $\mathcal{N}(\num{1e-3},\num{1e-4})$, respectively.
% \\
% \\
% \textcolor{red}{\textit{\textsc{\underline{DF}: text finish}}}
\subsection{Classical RNN Baseline}
We set the classical RNN, GRU and LSTM with the following model size to be the baseline in this study. The model sizes (number of parameters) are set to be similar to their quantum counterpart to investigate the learning capabilities of these models. For the experiments considered in this paper: RNN is with $40$ parameters in RNN and $6$ parameters in the final linear layer; GRU is with $120$ parameters in GRU and $6$ parameters in the final linear layer; LSTM is with $160$ parameters in LSTM and $6$ parameters in the final linear layer. Similar to the setting in quantum models, RC training means the recurrent parameters are frozen after randomly initialized and only final linear layers are trained.
\subsection{Tasks}
\subsubsection{\label{sec:function_approximation_damped_SHM}Function Approximation-Damped SHM}
Damped harmonic oscillators can be used to describe or approximate a wide range of systems, including the mass on a string and acoustic systems.  Damped harmonic oscillation can be described by the equation:
\begin{equation}
    \frac{\mathrm{d}^{2} x}{\mathrm{d} t^{2}}+2 \zeta \omega_{0} \frac{\mathrm{d} x}{\mathrm{d} t}+\omega_{0}^{2} x=0,
\end{equation}
where $\omega_{0}=\sqrt{\frac{k}{m}}$ is the (undamped) system's characteristic frequency and $\zeta=\frac{c}{2 \sqrt{m k}}$ is the damping ratio. In this paper, we consider a specific example from the simple pendulum with the following formulation:
\begin{equation}
\frac{d^{2} \theta}{d t^{2}}+\frac{b}{m} \frac{d \theta}{d t}+\frac{g}{L} \sin \theta=0,
\end{equation}
in which the gravitational constant $g = 9.81$, the damping factor $b = 0.15$, the pendulum length $l = 1$ and mass $m = 1$. The initial condition at $t = 0$ has angular displacement $\theta = 0$, and the angular velocity $\dot{\theta} = 3$ rad/sec.
We present the quantum learning result of the angular velocity $\dot{\theta}$.
\subsubsection{\label{sec:function_approximation_bessel}Function Approximation-Bessel Function}
Bessel functions are also commonly encountered in physics and engineering problems, such as electromagnetic fields or heat conduction in a cylindrical geometry.
Bessel functions of the first kind, $J_\alpha(x)$, are solutions to the Bessel differential equation
\begin{equation}
    x^{2} \frac{d^{2} y}{d x^{2}}+x \frac{d y}{d x}+\left(x^{2}-\alpha^{2}\right) y=0,
\end{equation}
and can be defined as 
\begin{equation}
    J_{\alpha}(x)=\sum_{m=0}^{\infty} \frac{(-1)^{m}}{m!  \Gamma(m+\alpha+1)}\left(\frac{x}{2}\right)^{2 m+\alpha},
\end{equation}
where $\Gamma(x)$ is the Gamma function.

In this paper, we choose $J_2$ as the function used for training.
\subsubsection{Time Series Prediction (NARMA Benchmark)}
\label{sec:time_series_prediction}
We use NARMA (Non-linear Auto-Regressive Moving Average) time series datasets \cite{NarmaAtiyaA} for this task. 
%
% The first NARMA example we consider is studied in \cite{} where the output sequence $\left\{y_{t}\right\}_{t=1}^{M}$ is described as 
% %
% \begin{equation}
% y_{t+1}=0.4 y_{t}+0.4 y_{t} y_{t-1}+0.6 u_{t}^{3}+0.1
% \end{equation}
% where $u_{t}$ and $y_{t}$ are the input and target sequences.
%
The NARMA series that we use in this work can be defined by \cite{NarmaGoudarzi}:
\begin{equation}
y_{t+1}=\alpha y_{t}+\beta y_{t}\left(\sum_{j=0}^{n_{o}-1} y_{t-j}\right)+\gamma u_{t-n_{o}+1} u_{t}+\delta
\end{equation}
where $(\alpha, \beta, \gamma, \delta)=(0.3,0.05,1.5,0.1)$ and $n_{0}$ is used to determine the nonlinearity.
The input $\left\{u_{t}\right\}_{t=1}^{M}$ for the NARMA tasks is:
\begin{equation}
u_{t}=0.1\left(\sin \left(\frac{2 \pi \bar{\alpha} t}{T}\right) \sin \left(\frac{2 \pi \bar{\beta} t}{T}\right) \sin \left(\frac{2 \pi \bar{\gamma} t}{T}\right)+1\right)
\end{equation}
where $(\bar{\alpha}, \bar{\beta}, \bar{\gamma}, T)=(2.11,3.73,4.11,100)$ as used in \cite{suzuki2022natural}. We set the length of inputs and outputs to $M = 300$. In this paper, we consider $n_{0} = 5$ and  $n_{0} = 10$, NARMA5 and NARMA10 respectively.
\section{\label{sec:Results}Results}
In the results we present here, the orange dashed line represents the ground truth while the blue solid line is the output from the models. The vertical red dashed line separates the \emph{training} set (left) from the \emph{testing} set (right). For all datasets we consider in this paper, $67\%$ are used for the training and the remaining $33\%$ are for testing.
%
% \YC{Do we need to compare with classical RNN, GRU, LSTM?}
% \YC{loss table to compare}
%

\subsection{Function Approximation}
\subsubsection{QRNN}
For the QRNN, we observe similar results in both the damped SHM (\figureautorefname{\ref{fig:rnn_damped_SHM}}) and Bessel function (\figureautorefname{\ref{fig:rnn_bessel}}) cases. Both the QRNN-RC and QRNN learn the important features after single training epochs. However, the fully trained QRNN captures more amplitude information in the first epoch. This is not surprising since the fully trained one requires more resources to tune all the quantum parameters, while the RC version does not. We observe that QRNN-RC can achieve performance comparable to fully trained QRNN after 15 epochs of training, except some of the large amplitude regions. After the training, the loss of RC and fully trained converge to a low value. 
% \YC{compare to classical}
%
In the case of damped SHM, if we compare the QRNN-RC to classical RNN-RC and RNN, we can observe that the QRNN-RC beats RNN-RC even after 100 epochs of training and reaches comparable performance to fully trained RNN. The results are similar in the case of Bessel function case, we observe that QRNN-RC beat RNN-RC from Epoch 1 to Epoch 100.
If we further add quantum device noises to the simulation (defined in \sectionautorefname{\ref{sec:noise_simulation_definition}}), we can observe that both the fully trained and RC QRNN reach pretty good performance after 100 epochs of training (shown in \figureautorefname{\ref{fig:noisy_rnn}}). Particularly, in both the damped SHM and Bessel function cases, we see that QRNN-RC can provide smoother outputs than the fully trained QRNN. We summarize the loss values of noise-free and noisy simulations in \tableautorefname{\ref{tab:noise_free_qrnn}} and \tableautorefname{\ref{tableNoisyMSE}} respectively.

\begin{table}[htbp]
\resizebox{1\textwidth}{!}{
\begin{tabular}{|l|l|l|l|l|l|l|}
\hline
Dataset    & Model & Reservoir & Epoch 1                                     & Epoch 15                                    & Epoch 30                                    & Epoch 100                                   \\ \hline
Damped SHM & QRNN  & True      & $1.72 \times 10^{-1}$/$2.21 \times 10^{-2}$ & $1.81 \times 10^{-2}$/$5.17 \times 10^{-3}$ & $1.81 \times 10^{-2}$/$4.96 \times 10^{-3}$ & $1.18 \times 10^{-2}$/$4.91 \times 10^{-3}$ \\ \hline
Damped SHM & QRNN  & False     & $1.01 \times 10^{-1}$/$5.04 \times 10^{-3}$ & $7.19 \times 10^{-3}$/$9.06 \times 10^{-4}$ & $1.87 \times 10^{-3}$/$2.16 \times 10^{-5}$ & $6.8 \times 10^{-4}$/$4.89 \times 10^{-5}$  \\ \hline
Damped SHM & RNN   & True      & $2.02 \times 10^{-1}$/$6.46 \times 10^{-2}$ & $1.29 \times 10^{-1}$/$2.68 \times 10^{-2}$ & $9.49 \times 10^{-2}$/$1.97 \times 10^{-2}$ & $2.87 \times 10^{-2}$/$5.87 \times 10^{-3}$ \\ \hline
Damped SHM & RNN   & False     & $7.22 \times 10^{-1}$/$6.78 \times 10^{-2}$ & $1.66 \times 10^{-2}$/$4.04 \times 10^{-3}$ & $2.82 \times 10^{-3}$/$9.40 \times 10^{-4}$ & $1.69 \times 10^{-3}$/$3.60 \times 10^{-4}$ \\ \hline
Bessel     & QRNN  & True      & $1.77 \times 10^{-1}$/$2.91 \times 10^{-2}$ & $1.65 \times 10^{-2}$/$3.60 \times 10^{-3}$ & $1.53 \times 10^{-2}$/$3.61 \times 10^{-3}$ & $1.52 \times 10^{-2}$/$3.57 \times 10^{-3}$ \\ \hline
Bessel     & QRNN  & False     & $4.16 \times 10^{-2}$/$5.54 \times 10^{-3}$ & $5.10 \times 10^{-3}$/$6.08 \times 10^{-4}$ & $1.40 \times 10^{-3}$/$3.19 \times 10^{-5}$ & $6.45 \times 10^{-4}$/$2.62 \times 10^{-5}$ \\ \hline
Bessel     & RNN   & True      & $5.22 \times 10^{-1}$/$1.65 \times 10^{-1}$ & $7.82 \times 10^{-2}$/$1.93 \times 10^{-2}$ & $7.11 \times 10^{-2}$/$1.76 \times 10^{-2}$ & $4.37 \times 10^{-2}$/$1.11 \times 10^{-2}$ \\ \hline
Bessel     & RNN   & False     & $1.79 \times 10^{-1}$/$2.93 \times 10^{-2}$ & $3.80 \times 10^{-3}$/$2.83 \times 10^{-4}$ & $4.49 \times 10^{-3}$/$3.25 \times 10^{-3}$ & $3.05 \times 10^{-4}$/$1.79 \times 10^{-5}$ \\ \hline
NARMA5     & QRNN  & True      & $2.72 \times 10^{-3}$/$2.93 \times 10^{-4}$ & $1.03 \times 10^{-4}$/$7.28 \times 10^{-5}$ & $1.26 \times 10^{-4}$/$3.44 \times 10^{-5}$ & $1.40 \times 10^{-4}$/$4.13 \times 10^{-5}$ \\ \hline
NARMA5     & QRNN  & False     & $3.19 \times 10^{-2}$/$1.82 \times 10^{-4}$ & $3.26 \times 10^{-4}$/$9.52 \times 10^{-5}$ & $1.57 \times 10^{-4}$/$4.64 \times 10^{-5}$ & $1.84 \times 10^{-4}$/$4.63 \times 10^{-5}$ \\ \hline
NARMA5     & RNN   & True      & $2.73 \times 10^{-2}$/$7.02 \times 10^{-3}$ & $1.78 \times 10^{-4}$/$7.28 \times 10^{-5}$ & $1.73 \times 10^{-4}$/$7.07 \times 10^{-5}$ & $1.45 \times 10^{-4}$/$6.01 \times 10^{-5}$ \\ \hline
NARMA5     & RNN   & False     & $1.03 \times 10^{-1}$/$2.94 \times 10^{-2}$ & $3.47 \times 10^{-4}$/$1.37 \times 10^{-4}$ & $3.27 \times 10^{-4}$/$1.29 \times 10^{-4}$ & $2.44 \times 10^{-4}$/$9.76 \times 10^{-5}$ \\ \hline
NARMA10    & QRNN  & True      & $1.00 \times 10^{-1}$/$8.73 \times 10^{-3}$ & $2.22 \times 10^{-4}$/$7.99 \times 10^{-5}$ & $2.63 \times 10^{-4}$/$9.42 \times 10^{-5}$ & $3.03 \times 10^{-4}$/$1.24 \times 10^{-4}$ \\ \hline
NARMA10    & QRNN  & False     & $3.54 \times 10^{-2}$/$1.03 \times 10^{-4}$ & $3.27 \times 10^{-4}$/$1.96 \times 10^{-4}$ & $3.65 \times 10^{-4}$/$1.55 \times 10^{-4}$ & $3.99 \times 10^{-4}$/$1.56 \times 10^{-4}$ \\ \hline
NARMA10    & RNN   & True      & $6.87 \times 10^{-2}$/$6.73 \times 10^{-4}$ & $5.82 \times 10^{-4}$/$1.75 \times 10^{-4}$ & $5.80 \times 10^{-4}$/$1.74 \times 10^{-4}$ & $5.70 \times 10^{-4}$/$1.71 \times 10^{-4}$ \\ \hline
NARMA10    & RNN   & False     & $2.70 \times 10^{-1}$/$5.75 \times 10^{-2}$ & $5.25 \times 10^{-4}$/$1.57 \times 10^{-4}$ & $5.07 \times 10^{-4}$/$1.51 \times 10^{-4}$ & $4.23 \times 10^{-4}$/$1.27 \times 10^{-4}$ \\ \hline
\end{tabular}}

\caption{RNN model results for training Epochs 1, 15, 30 and 100.}
\label{tab:noise_free_qrnn}
\end{table}

\begin{figure}[hbtp]
\includegraphics[width=1.\linewidth]{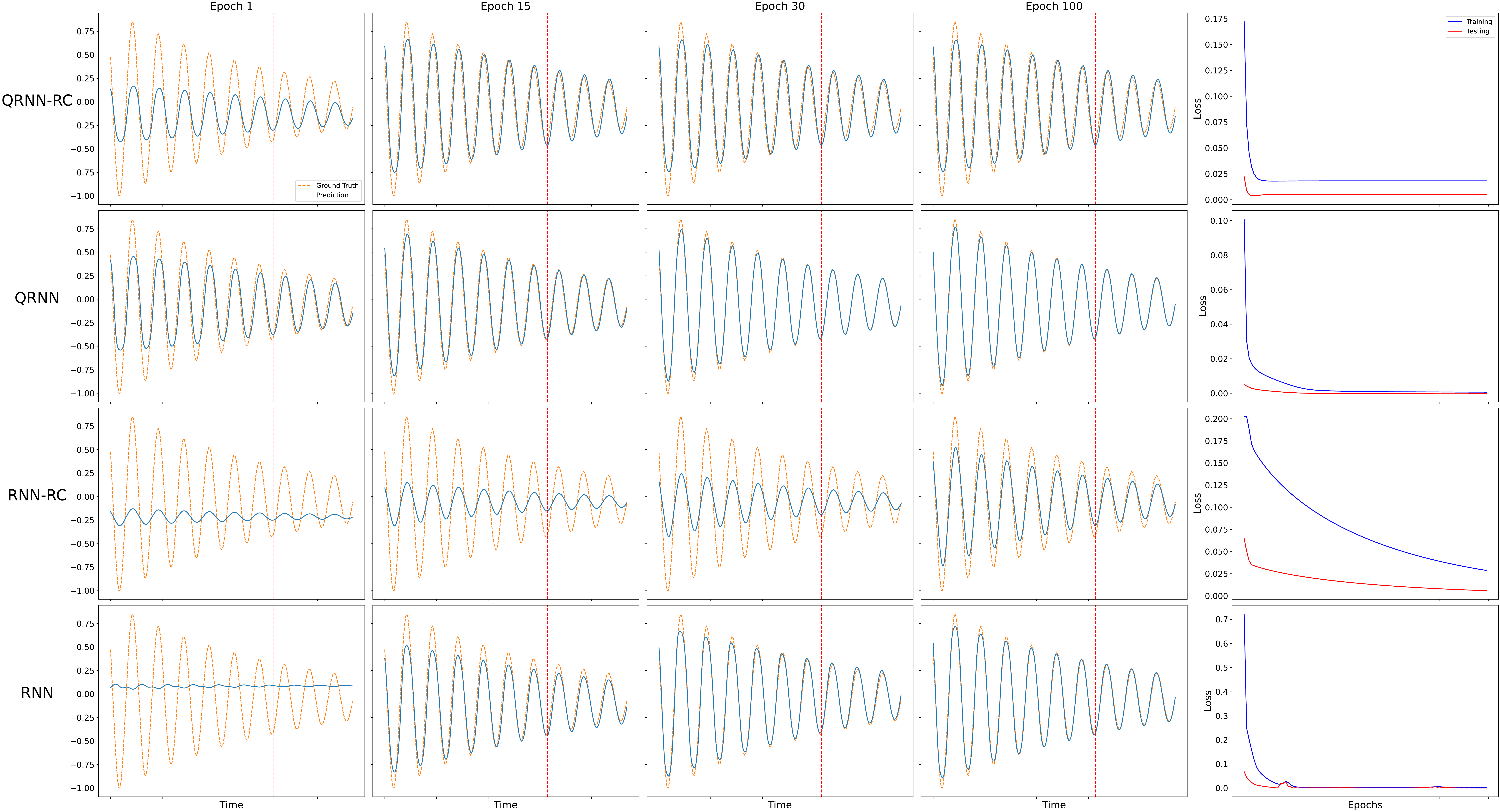}% Here is how to import EPS art
\caption{{\bfseries Learning the damped SHM with QRNN-RC.} 
%
% In the first epoch of training, we can observe that the fully trained QRNN learns more amplitude information than the QRNN-RC which only the final linear layer is trained. However, we observe that the QRNN-RC can reach comparable performance to QRNN after 15 epochs of training. If we compare the QRNN-RC to classical RNN-RC and RNN, we can observe that the QRNN-RC beats RNN-RC even after 100 epochs of training and reach comparable performance to fully trained RNN.
% %
% The orange dashed line represents the ground truth $\dot{\theta}$ while the blue solid line is the output from the models. The vertical red dashed line separates the \emph{training} set (left) from the \emph{testing} set (right).
}
\label{fig:rnn_damped_SHM}
\end{figure}
\begin{figure}[hbtp]
\includegraphics[width=1.\linewidth]{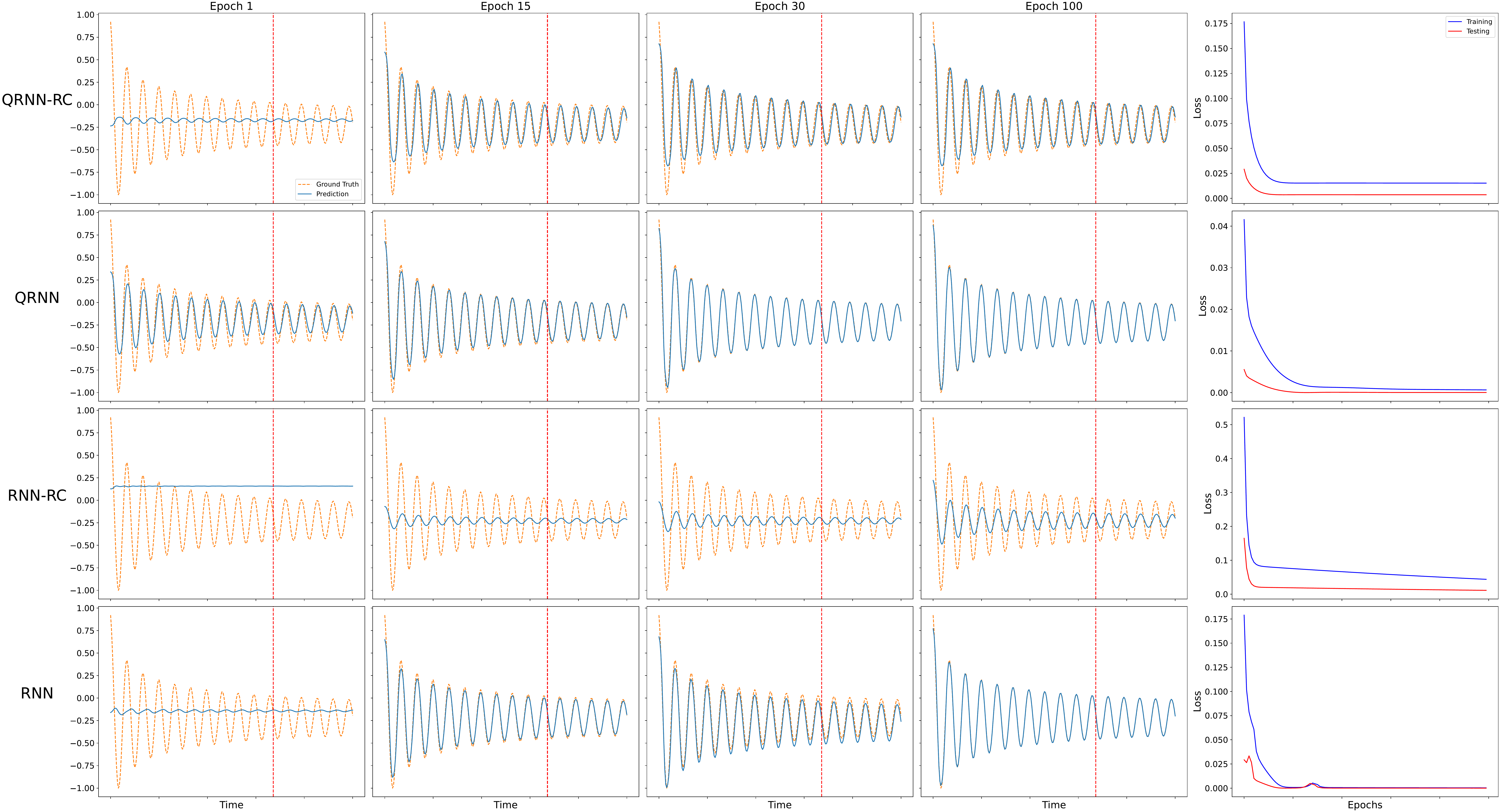}% Here is how to import EPS art
\caption{{\bfseries Learning the Bessel function with QRNN-RC.} 
%
% \YC{Continue here...}
%
% In the first epoch of training, we can observe that the fully trained QRNN learns more amplitude information than the QRNN-RC which only the final linear layer is trained. This result is consistent with what we found in the damped SHM (see \figureautorefname{\ref{fig:rnn_damped_SHM}}). We observe that QRNN-RC can reach performance comparable to fully trained QRNN after 15 epochs of training, except some of the large amplitude regions. 
% %
% If we compare the QRNN-RC to classical RNN-RC and RNN, we observe that QRNN-RC beat RNN-RC from Epoch 1 to Epoch 100 and can reach comparable performance to RNN after 15 epochs of training.
% %
% The orange dashed line represents the ground truth $J_2$ while the blue solid line is the output from the models. The vertical red dashed line separates the \emph{training} set (left) from the \emph{testing} set (right).
}
\label{fig:rnn_bessel}
\end{figure}
\begin{figure}[hbtp]
\includegraphics[width=1.\linewidth]{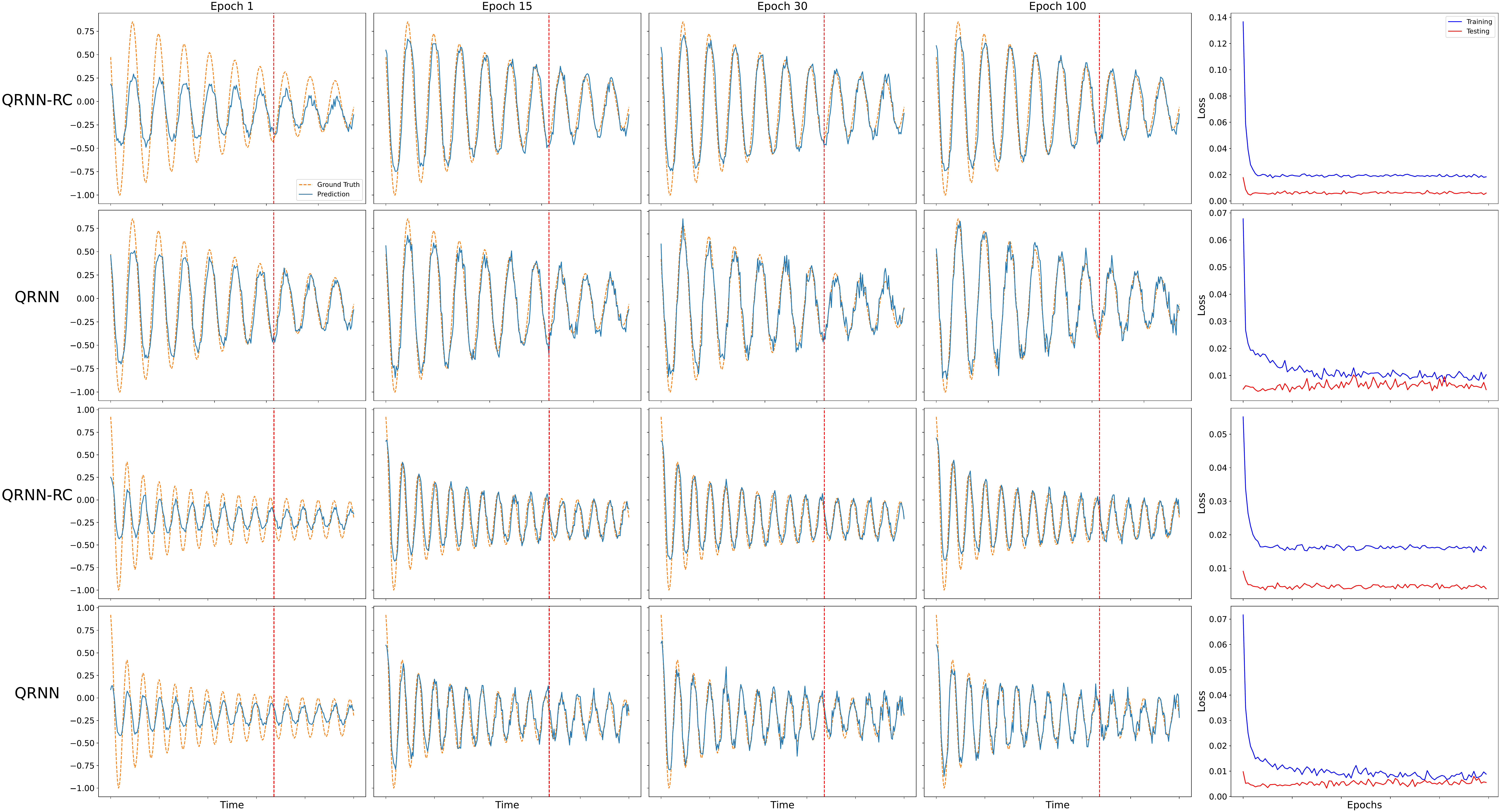}% Here is how to import EPS art
\caption{{\bfseries Noisy simulation of QRNN.} 
%
% We observe that the both the full optimization and RC training of QRNN under the effect of simulated quantum noises can still reach reasonable performance in both the damped SHM and the Bessel function.  
% %
% Importantly, we observe that the in both the damped SHM and Bessel function cases, the QRNN-RC can generate smoother outputs than the fully optimized QRNN. 
%
}
\label{fig:noisy_rnn}
\end{figure}

\subsubsection{QGRU}
For the QGRU, we observe similar results for both the damped SHM (\figureautorefname{\ref{fig:gru_damped_SHM}}) and Bessel function (\figureautorefname{\ref{fig:gru_bessel}}) cases.
After the first epoch of training, we can observe that the fully trained QGRU learns more amplitude information than the QGRU-RC, in which only the final linear layer is trained. In the case of damped SHM, we observe that the QGRU-RC can reach comparable performance to QGRU after 15 epochs of training. If we compare the QGRU-RC to classical GRU-RC and GRU, we can observe that the QGRU-RC beats GRU-RC up to the first 30 epochs of training and reaches similar performance to GRU-RC and fully trained GRU after 100 epochs of training. In the case of Bessel function, we observe that QGRU-RC saturates after 15 epochs of training and can capture most of the data, except some of the large amplitude regions. We also observe that the QGRU-RC performs similar to the classical GRU-RC after 15 epochs of training.

If we add quantum device noise to the simulation (defined in \sectionautorefname{\ref{sec:noise_simulation_definition}}), we observe that both the full optimization and RC training of QGRU under the effect of simulated quantum noises can still reach reasonable performance in both the damped SHM and the Bessel function (shown in \figureautorefname{\ref{fig:noisy_gru}}).  
Most importantly, we observe that the in both the damped SHM and Bessel function cases, the QGRU-RC can generate smoother outputs than the fully optimized QGRU. We summarize the loss values of noise-free and noisy simulations in \tableautorefname{\ref{tab:noise_free_qgru}} and \tableautorefname{\ref{tableNoisyMSE}} respectively.
\begin{table}[htbp]
\resizebox{1\linewidth}{!}{
\begin{tabular}{|l|l|l|l|l|l|l|}
\hline
Dataset    & Model & Reservoir & Epoch 1                                     & Epoch 15                                    & Epoch 30                                    & Epoch 100                                   \\ \hline
Damped SHM & QGRU  & True      & $2.26 \times 10^{-1}$/$2.78 \times 10^{-2}$ & $4.54 \times 10^{-2}$/$1.31 \times 10^{-2}$ & $4.55 \times 10^{-2}$/$1.30 \times 10^{-2}$ & $4.55 \times 10^{-2}$/$1.29 \times 10^{-2}$ \\ \hline
Damped SHM & QGRU  & False     & $1.97 \times 10^{-1}$/$1.51 \times 10^{-2}$ & $2.01 \times 10^{-2}$/$3.64 \times 10^{-3}$ & $1.04 \times 10^{-2}$/$1.30 \times 10^{-3}$ & $1.39 \times 10^{-3}$/$1.19 \times 10^{-4}$ \\ \hline
Damped SHM & GRU   & True      & $4.62 \times 10^{-1}$/$1.18 \times 10^{-1}$ & $1.13 \times 10^{-1}$/$2.26 \times 10^{-2}$ & $7.45 \times 10^{-2}$/$1.50 \times 10^{-2}$ & $4.61 \times 10^{-2}$/$9.92 \times 10^{-3}$ \\ \hline
Damped SHM & GRU   & False     & $2.12 \times 10^{-1}$/$8.54 \times 10^{-2}$ & $2.22 \times 10^{-2}$/$3.90 \times 10^{-3}$ & $2.94 \times 10^{-3}$/$1.80 \times 10^{-4}$ & $4.51 \times 10^{-4}$/$7.75 \times 10^{-5}$ \\ \hline
Bessel     & QGRU  & True      & $1.54 \times 10^{-1}$/$2.58 \times 10^{-2}$ & $3.90 \times 10^{-2}$/$9.92 \times 10^{-3}$ & $3.82 \times 10^{-2}$/$9.90 \times 10^{-3}$ & $3.82 \times 10^{-2}$/$9.89 \times 10^{-3}$ \\ \hline
Bessel     & QGRU  & False     & $5.53 \times 10^{-2}$/$9.39 \times 10^{-3}$ & $1.10 \times 10^{-2}$/$2.05 \times 10^{-3}$ & $2.94 \times 10^{-3}$/$9.69 \times 10^{-5}$ & $1.31 \times 10^{-3}$/$1.37 \times 10^{-5}$ \\ \hline
Bessel     & GRU   & True      & $1.71 \times 10^{-1}$/$3.33 \times 10^{-2}$ & $4.16 \times 10^{-2}$/$1.13 \times 10^{-2}$ & $3.78 \times 10^{-2}$/$1.05 \times 10^{-2}$ & $3.05 \times 10^{-2}$/$8.51 \times 10^{-3}$ \\ \hline
Bessel     & GRU   & False     & $1.03 \times 10^{-1}$/$9.98 \times 10^{-2}$ & $1.98 \times 10^{-2}$/$4.77 \times 10^{-3}$ & $4.62 \times 10^{-3}$/$1.72 \times 10^{-3}$ & $4.68 \times 10^{-4}$/$9.09 \times 10^{-6}$ \\ \hline
NARMA5     & QGRU  & True      & $9.48 \times 10^{-2}$/$6.59 \times 10^{-3}$ & $6.46 \times 10^{-5}$/$3.21 \times 10^{-5}$ & $8.62 \times 10^{-5}$/$2.36 \times 10^{-5}$ & $1.10 \times 10^{-4}$/$3.50 \times 10^{-5}$ \\ \hline
NARMA5     & QGRU  & False     & $4.12 \times 10^{-3}$/$5.58 \times 10^{-5}$ & $1.53 \times 10^{-4}$/$3.22 \times 10^{-5}$ & $1.36 \times 10^{-4}$/$2.41 \times 10^{-5}$ & $1.22 \times 10^{-4}$/$3.55 \times 10^{-5}$ \\ \hline
NARMA5     & GRU   & True      & $1.90 \times 10^{-3}$/$2.23 \times 10^{-2}$ & $3.62 \times 10^{-4}$/$1.45 \times 10^{-4}$ & $3.46 \times 10^{-4}$/$1.39 \times 10^{-4}$ & $2.69 \times 10^{-4}$/$1.09 \times 10^{-4}$ \\ \hline
NARMA5     & GRU   & False     & $9.00 \times 10^{-2}$/$6.43 \times 10^{-4}$ & $2.63 \times 10^{-4}$/$1.04 \times 10^{-4}$ & $2.36 \times 10^{-4}$/$9.39 \times 10^{-5}$ & $1.22 \times 10^{-4}$/$4.99 \times 10^{-5}$ \\ \hline
NARMA10    & QGRU  & True      & $1.54 \times 10^{-3}$/$2.91 \times 10^{-4}$ & $2.20 \times 10^{-4}$/$7.22 \times 10^{-5}$ & $2.50 \times 10^{-4}$/$9.23 \times 10^{-5}$ & $2.74 \times 10^{-4}$/$1.21 \times 10^{-4}$ \\ \hline
NARMA10    & QGRU  & False     & $6.30 \times 10^{-3}$/$1.28 \times 10^{-4}$ & $3.63 \times 10^{-4}$/$1.14 \times 10^{-4}$ & $4.04 \times 10^{-4}$/$1.20 \times 10^{-4}$ & $2.97 \times 10^{-4}$/$1.25 \times 10^{-4}$ \\ \hline
NARMA10    & GRU   & True      & $2.08 \times 10^{-1}$/$8.04 \times 10^{-2}$ & $2.25 \times 10^{-4}$/$8.21 \times 10^{-5}$ & $2.18 \times 10^{-4}$/$7.57 \times 10^{-5}$ & $2.14 \times 10^{-4}$/$7.50 \times 10^{-5}$ \\ \hline
NARMA10    & GRU   & False     & $5.39 \times 10^{-1}$/$6.93 \times 10^{-2}$ & $2.56 \times 10^{-4}$/$8.03 \times 10^{-5}$ & $2.52 \times 10^{-4}$/$7.94 \times 10^{-5}$ & $2.32 \times 10^{-4}$/$7.47 \times 10^{-5}$ \\ \hline
\end{tabular}
}
\caption{GRU model results for training Epochs 1, 15, 30 and 100.}
\label{tab:noise_free_qgru}
\end{table}
\begin{figure}[hbtp]
\includegraphics[width=1.\linewidth]{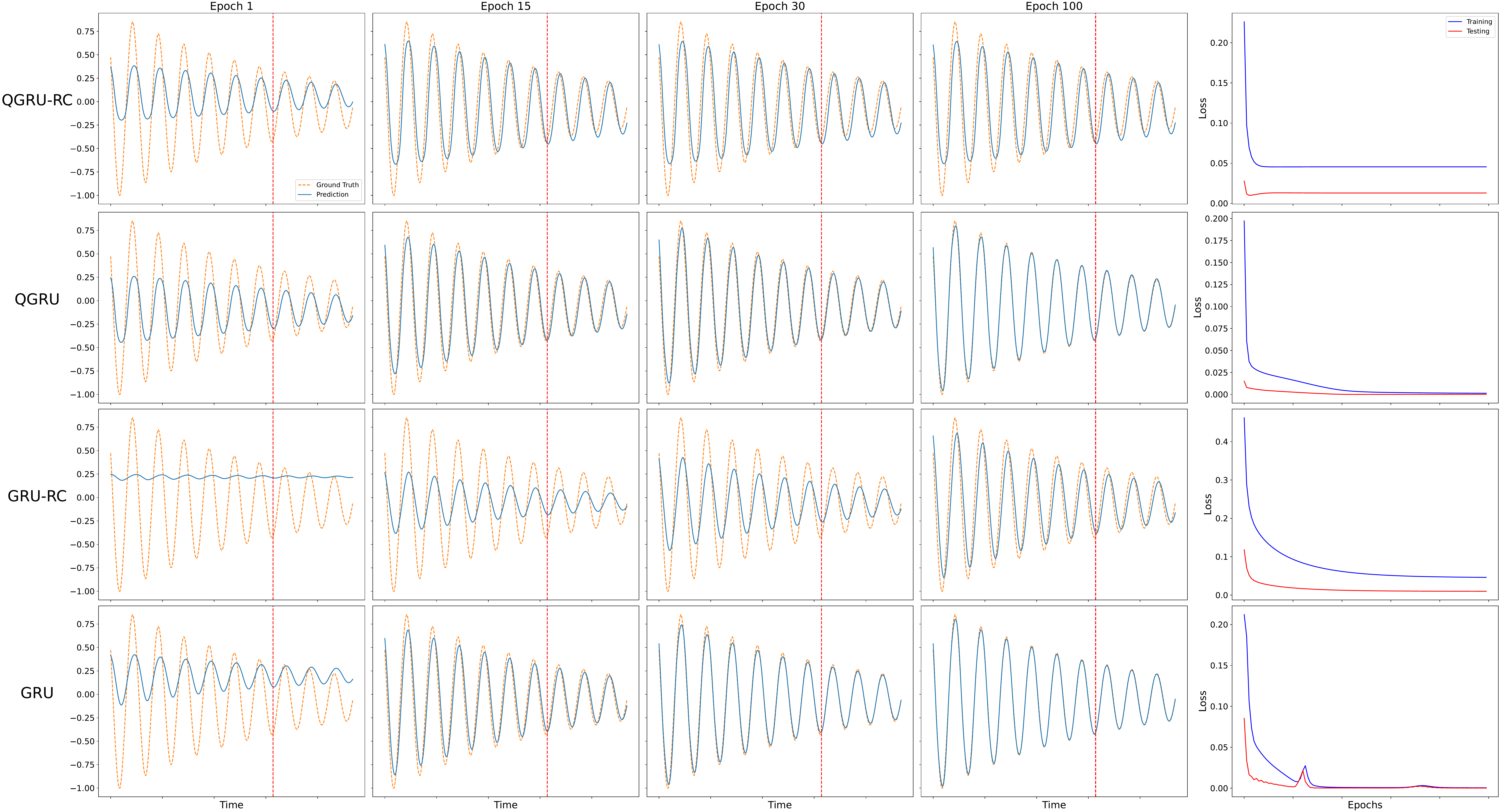}% Here is how to import EPS art
\caption{{\bfseries Learning the damped SHM with QGRU-RC.} 
% In the first epoch of training, we can observe that the fully trained QGRU learns more amplitude information than the QGRU-RC which only the final linear layer is trained. However, we observe that the QGRU-RC can reach comparable performance to QGRU after 15 epochs of training. If we compare the QGRU-RC to classical GRU-RC and GRU, we can observe that the QGRU-RC beats GRU-RC up to the first 30 epochs of training and reach similar performance to GRU-RC and fully trained GRU after 100 epochs of training.
% %
% The orange dashed line represents the ground truth $\dot{\theta}$ while the blue solid line is the output from the models. The vertical red dashed line separates the \emph{training} set (left) from the \emph{testing} set (right).
}
\label{fig:gru_damped_SHM}
\end{figure}
\begin{figure}[hbtp]
\includegraphics[width=1.\linewidth]{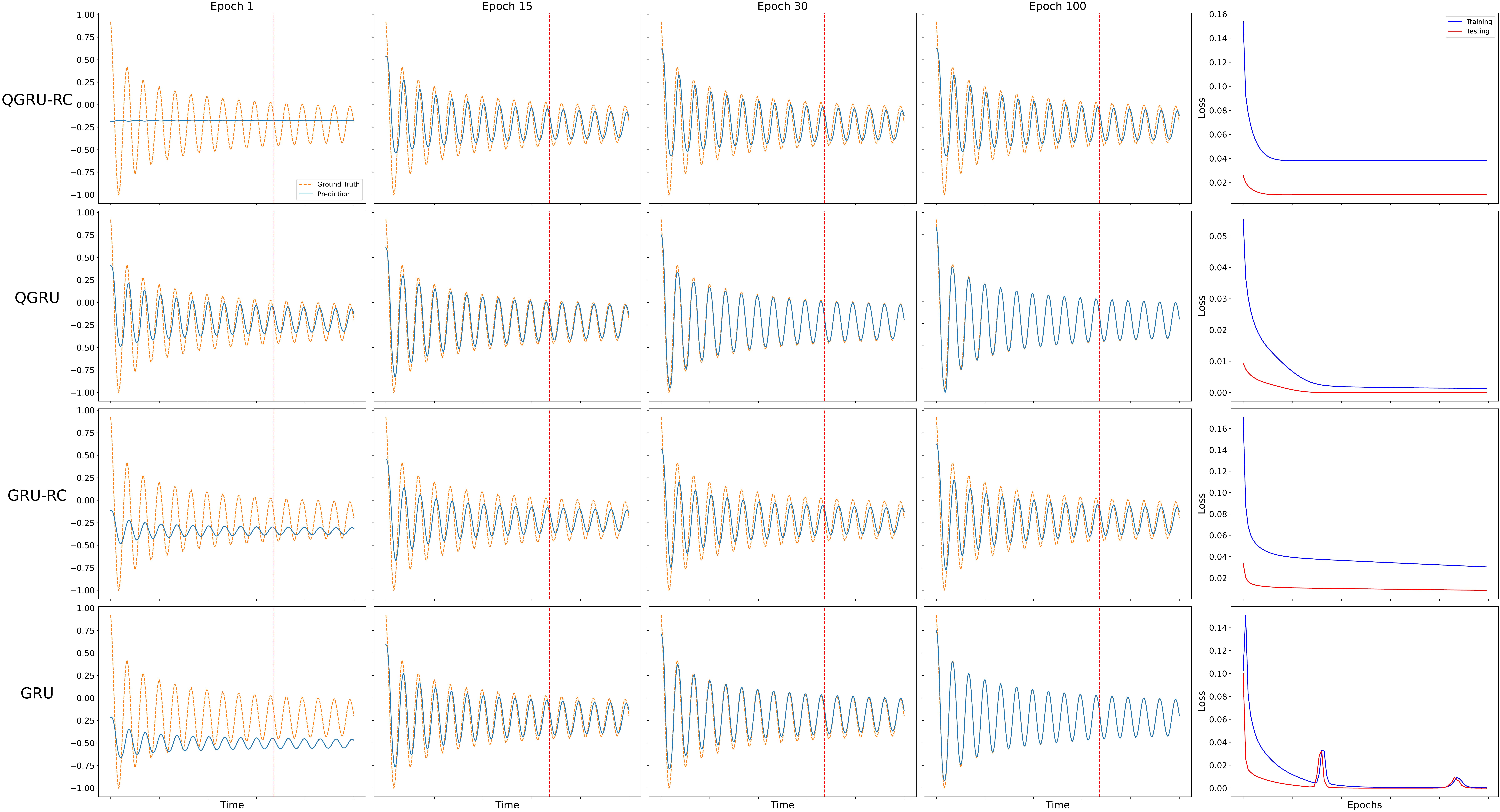}% Here is how to import EPS art
\caption{{\bfseries Learning the Bessel function with QGRU-RC.}
%
% In the first epoch of training, we can observe that the fully trained QGRU learns more amplitude information than the QGRU-RC which only the final linear layer is trained. This result is consistent with what we found in the damped SHM (see \figureautorefname{\ref{fig:gru_damped_SHM}}). We observe that QGRU-RC saturates after 15 epochs of training and can capture most part of the data, except some of the large amplitude regions.
% %
% We also observe that the QGRU-RC performs similar to the classical GRU-RC after 15 epochs of training.
% %
% The orange dashed line represents the ground truth $J_2$ while the blue solid line is the output from the models. The vertical red dashed line separates the \emph{training} set (left) from the \emph{testing} set (right).
}
\label{fig:gru_bessel}
\end{figure}
\begin{figure}[hbtp]
\includegraphics[width=1.\linewidth]{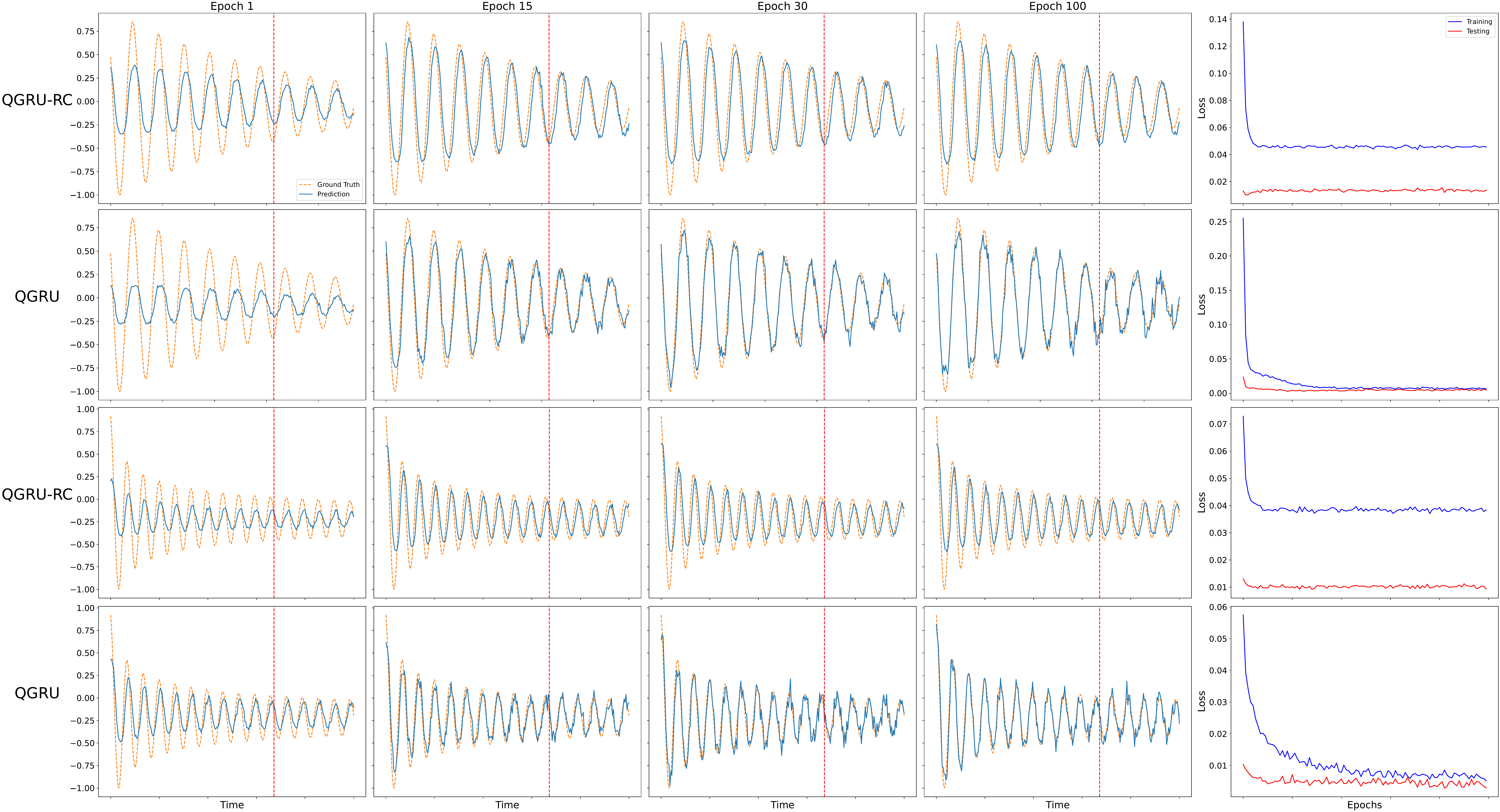}% Here is how to import EPS art
\caption{{\bfseries Noisy simulation of QGRU.}
% We observe that the both the full optimization and RC training of QGRU under the effect of simulated quantum noises can still reach reasonable performance in both the damped SHM and the Bessel function.  
% %
% Importantly, we observe that the in both the damped SHM and Bessel function cases, the QGRU-RC can generate smoother outputs than the fully optimized QGRU. 
}
\label{fig:noisy_gru}
\end{figure}
\subsubsection{QLSTM}
For the QLSTM, we observe similar results in both the damped SHM (\figureautorefname{\ref{fig:lstm_damped_SHM}}) and Bessel function (\figureautorefname{\ref{fig:lstm_bessel}}) cases. For the damped SHM case, we observe that after the first epoch of training, the fully trained QLSTM learns more amplitude information than the QLSTM-RC in which only the final linear layer is trained. While in the Bessel function case, the QLSTM-RC and QLSTM provide similar learning outcomes in the first training epoch. We observe that both models reach similar results after 100 epochs of training. However, the loss values of QLSTM are much lower after the training. This is not surprising since all the model parameters are trained in QLSTM while in QLSTM-RC only the final linear layer is trained.
If we compare QLSTM-RC to LSTM-RC, we can observe that the quantum version captures more features after the same number of training epochs in both the damped SHM and Bessel function cases.
If we add quantum device noise to the simulation (defined in \sectionautorefname{\ref{sec:noise_simulation_definition}}), we observe that the both the full optimization and RC training of QLSTM under the effect of simulated quantum noise can still reach reasonable performance in both the damped SHM and the Bessel function (shown in \figureautorefname{\ref{fig:noisy_lstm}}).  
We observe that the in both the damped SHM and Bessel function cases, the QLSTM-RC can generate smoother outputs than the fully optimized QLSTM. The results are consistent with QRNN and QGRU. We summarize the loss values of noise-free and noisy simulations in \tableautorefname{\ref{tab:noise_free_qlstm}} and \tableautorefname{\ref{tableNoisyMSE}} respectively.
\begin{table}[htbp]
\resizebox{1\linewidth}{!}{
\begin{tabular}{|l|l|l|l|l|l|l|}
\hline
Dataset    & Model & Reservoir & Epoch 1                                     & Epoch 15                                    & Epoch 30                                    & Epoch 100                                   \\ \hline
Damped SHM & QLSTM & True      & $3.19 \times 10^{-1}$/$5.86 \times 10^{-2}$ & $6.42 \times 10^{-2}$/$1.08 \times 10^{-2}$ & $5.55 \times 10^{-2}$/$1.38 \times 10^{-2}$ & $5.55 \times 10^{-2}$/$1.41 \times 10^{-2}$ \\ \hline
Damped SHM & QLSTM & False     & $1.66 \times 10^{-1}$/$1.35 \times 10^{-2}$ & $2.89 \times 10^{-2}$/$5.53 \times 10^{-3}$ & $9.06 \times 10^{-3}$/$3.41 \times 10^{-4}$ & $2.86 \times 10^{-3}$/$1.94 \times 10^{-4}$ \\ \hline
Damped SHM & LSTM  & True      & $3.45 \times 10^{-1}$/$7.49 \times 10^{-2}$ & $1.89 \times 10^{-1}$/$3.98 \times 10^{-2}$ & $1.66 \times 10^{-1}$/$3.51 \times 10^{-2}$ & $1.10 \times 10^{-1}$/$2.32 \times 10^{-2}$ \\ \hline
Damped SHM & LSTM  & False     & $3.32 \times 10^{-1}$/$3.29 \times 10^{-2}$ & $3.65 \times 10^{-2}$/$7.38 \times 10^{-3}$ & $6.74 \times 10^{-3}$/$7.27 \times 10^{-4}$ & $2.32 \times 10^{-3}$/$1.68 \times 10^{-3}$ \\ \hline
Bessel     & QLSTM & True      & $7.53 \times 10^{-2}$/$1.36 \times 10^{-2}$ & $3.94 \times 10^{-2}$/$9.67 \times 10^{-3}$ & $3.90 \times 10^{-2}$/$1.01 \times 10^{-2}$ & $3.90 \times 10^{-2}$/$1.02 \times 10^{-2}$ \\ \hline
Bessel     & QLSTM & False     & $1.04 \times 10^{-1}$/$1.66 \times 10^{-2}$ & $2.30 \times 10^{-2}$/$5.35 \times 10^{-3}$ & $1.27 \times 10^{-2}$/$2.42 \times 10^{-3}$ & $6.97 \times 10^{-4}$/$1.21 \times 10^{-5}$ \\ \hline
Bessel     & LSTM  & True      & $1.21 \times 10^{-1}$/$2.46 \times 10^{-2}$ & $6.58 \times 10^{-2}$/$1.65 \times 10^{-2}$ & $5.43 \times 10^{-2}$/$1.39 \times 10^{-2}$ & $3.76 \times 10^{-2}$/$1.02 \times 10^{-2}$ \\ \hline
Bessel     & LSTM  & False     & $3.03 \times 10^{-1}$/$4.55 \times 10^{-2}$ & $3.48 \times 10^{-2}$/$8.71 \times 10^{-3}$ & $6.97 \times 10^{-3}$/$1.41 \times 10^{-3}$ & $1.31 \times 10^{-3}$/$3.53 \times 10^{-4}$ \\ \hline
NARMA5     & QLSTM & True      & $8.54 \times 10^{-4}$/$5.40 \times 10^{-4}$ & $1.32 \times 10^{-4}$/$1.10 \times 10^{-4}$ & $9.06 \times 10^{-5}$/$2.96 \times 10^{-5}$ & $1.13 \times 10^{-4}$/$2.58 \times 10^{-5}$ \\ \hline
NARMA5     & QLSTM & False     & $3.99 \times 10^{-3}$/$4.07 \times 10^{-4}$ & $3.30 \times 10^{-4}$/$4.23 \times 10^{-4}$ & $1.86 \times 10^{-4}$/$2.06 \times 10^{-4}$ & $9.85 \times 10^{-5}$/$2.52 \times 10^{-5}$ \\ \hline
NARMA5     & LSTM  & True      & $4.15 \times 10^{-2}$/$2.10 \times 10^{-4}$ & $3.73 \times 10^{-4}$/$1.48 \times 10^{-4}$ & $3.72 \times 10^{-4}$/$1.48 \times 10^{-4}$ & $3.65 \times 10^{-4}$/$1.45 \times 10^{-4}$ \\ \hline
NARMA5     & LSTM  & False     & $1.19 \times 10^{-1}$/$7.97 \times 10^{-4}$ & $3.34 \times 10^{-4}$/$1.38 \times 10^{-4}$ & $2.93 \times 10^{-4}$/$1.15 \times 10^{-4}$ & $1.91 \times 10^{-4}$/$8.78 \times 10^{-5}$ \\ \hline
NARMA10    & QLSTM & True      & $1.97 \times 10^{-3}$/$2.78 \times 10^{-4}$ & $3.01 \times 10^{-4}$/$1.39 \times 10^{-4}$ & $2.36 \times 10^{-4}$/$8.78 \times 10^{-5}$ & $2.59 \times 10^{-4}$/$9.64 \times 10^{-5}$ \\ \hline
NARMA10    & QLSTM & False     & $4.19 \times 10^{-3}$/$4.71 \times 10^{-4}$ & $3.35 \times 10^{-4}$/$4.73 \times 10^{-4}$ & $3.20 \times 10^{-4}$/$3.74 \times 10^{-4}$ & $2.59 \times 10^{-4}$/$9.50 \times 10^{-5}$ \\ \hline
NARMA10    & LSTM  & True      & $1.16 \times 10^{-2}$/$4.50 \times 10^{-3}$ & $4.26 \times 10^{-4}$/$1.27 \times 10^{-4}$ & $4.17 \times 10^{-4}$/$1.25 \times 10^{-4}$ & $3.74 \times 10^{-4}$/$1.12 \times 10^{-4}$ \\ \hline
NARMA10    & LSTM  & False     & $1.70 \times 10^{-1}$/$4.21 \times 10^{-4}$ & $2.94 \times 10^{-4}$/$8.68 \times 10^{-5}$ & $2.76 \times 10^{-4}$/$8.53 \times 10^{-5}$ & $2.31 \times 10^{-4}$/$8.12 \times 10^{-5}$ \\ \hline
\end{tabular}}
\caption{LSTM model results for training Epochs 1, 15, 30 and 100.}
\label{tab:noise_free_qlstm}
\end{table}
\begin{figure}[hbtp]
\includegraphics[width=1.\linewidth]{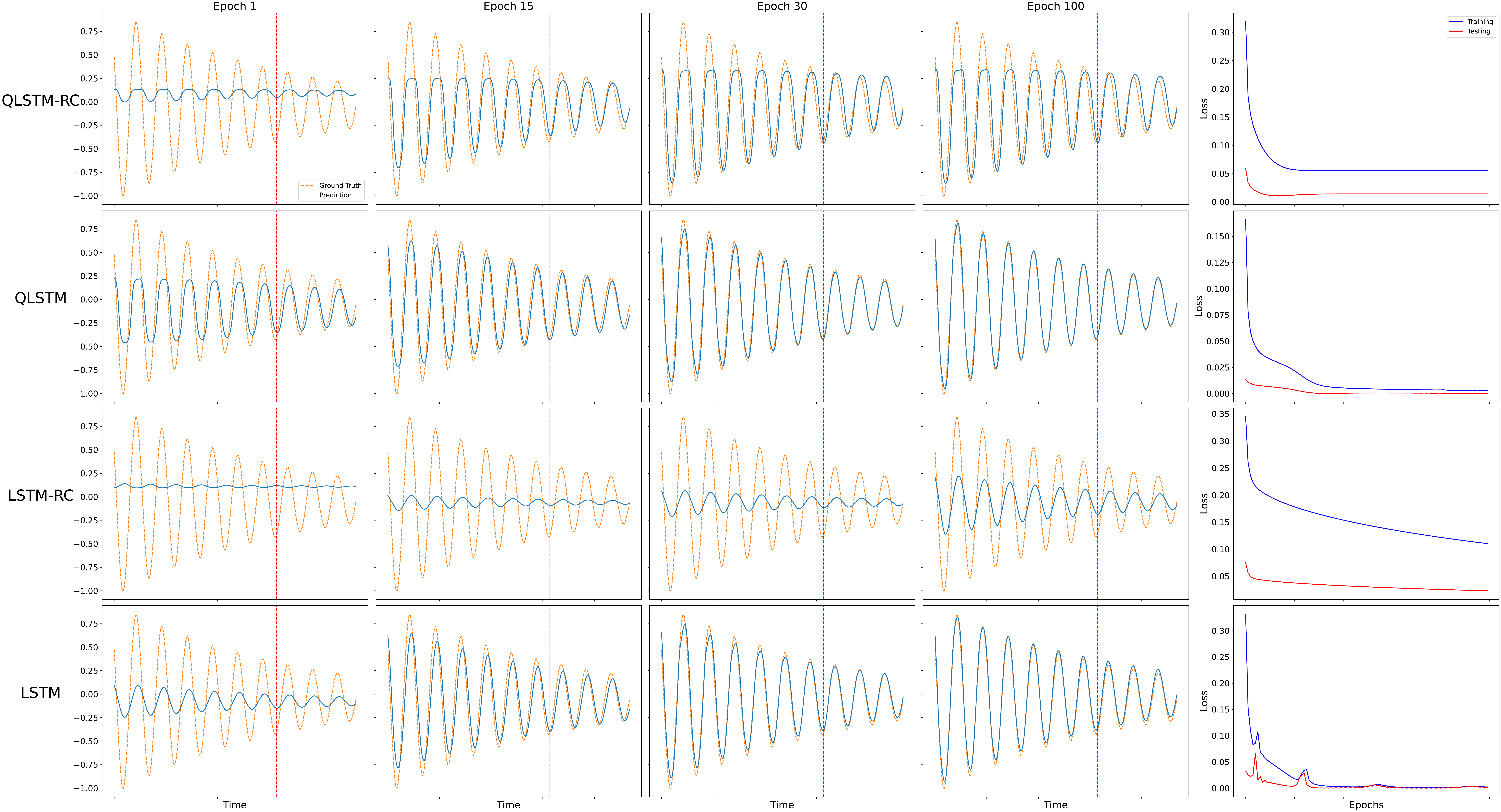}% Here is how to import EPS art
\caption{{\bfseries Learning the damped SHM with QLSTM-RC.} 
% In the first epoch of training, we can observe that the fully trained QLSTM learns more amplitude information than the QLSTM-RC which only the final linear layer is trained. 
%
% If we compare QLSTM-RC to LSTM-RC, we can observe that the quantum version capture more features after same number of training epochs.
%
% The orange dashed line represents the ground truth $\dot{\theta}$ while the blue solid line is the output from the models. The vertical red dashed line separates the \emph{training} set (left) from the \emph{testing} set (right).
}
\label{fig:lstm_damped_SHM}
\end{figure}
\begin{figure}[hbtp]
\includegraphics[width=1.\linewidth]{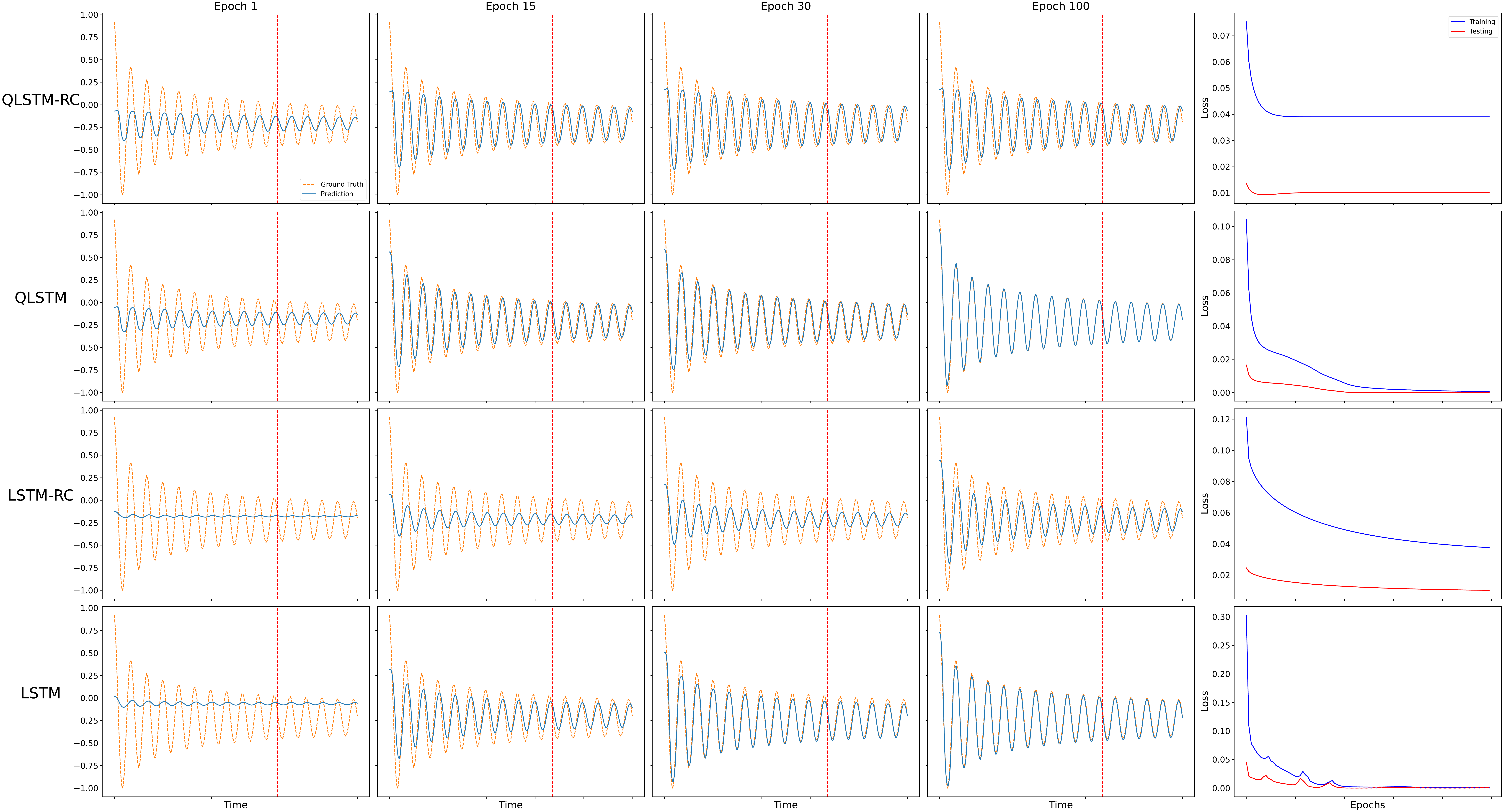}% Here is how to import EPS art
\caption{{\bfseries Learning the Bessel function with QLSTM-RC.}
%
% The QLSTM-RC and QLSTM provide similar learning outcomes in the first training epoch. We observe that both models reach similar results after 100 epochs of training. However, the loss values of QLSTM are much lower after the training. This is not surprising since all the model parameters are trained in QLSTM while in QLSTM-RC only the final linear layer is trained.
%
% We also observe that the QLSTM-RC beats LSTM-RC in all training epochs. 
%
% The orange dashed line represents the ground truth $J_2$ while the blue solid line is the output from the models. The vertical red dashed line separates the \emph{training} set (left) from the \emph{testing} set (right).
}
\label{fig:lstm_bessel}
\end{figure}
\begin{figure}[hbtp]
\includegraphics[width=1.\linewidth]{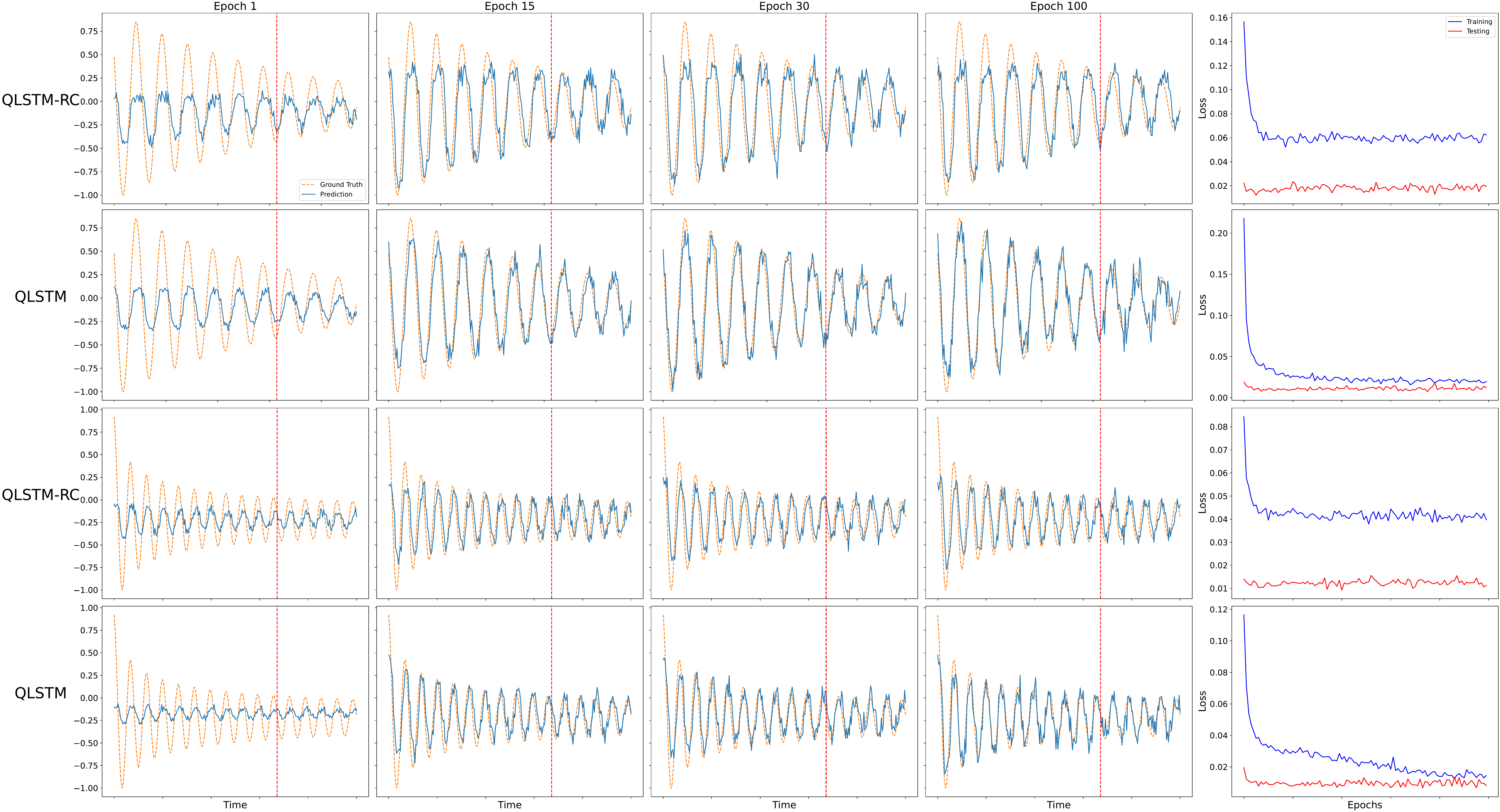}% Here is how to import EPS art
\caption{{\bfseries Noisy simulation of QLSTM.}
% We observe that the both the full optimization and RC training of QLSTM under the effect of simulated quantum noises can still reach reasonable performance in both the damped SHM and the Bessel function.  
%
% Importantly, we observe that the in both the damped SHM and Bessel function cases, the QLSTM-RC can generate smoother outputs than the fully optimized QLSTM. 
}
\label{fig:noisy_lstm}
\end{figure}
\begin{table}[htbp]
% \centering
\resizebox{1\linewidth}{!}{
\begin{tabular}{|l|l|l|l|l|l|l|}
\hline
Data & Model & Reservoir & Epoch 1 & Epoch 15 & Epoch 30 & Epoch 100 \\
\hline
Bessel & GRU & False & $3.88 \times 10^{-2}$/$9.71 \times 10^{-3}$ & $1.33 \times 10^{-2}$/$4.7 \times 10^{-3}$ & $9.65 \times 10^{-3}$/$5.32 \times 10^{-3}$ & $5.43 \times 10^{-3}$/$4.47 \times 10^{-3}$ \\
\cline{1-7}
Bessel & GRU & True & $5.18 \times 10^{-2}$/$1.3 \times 10^{-2}$ & $3.72 \times 10^{-2}$/$1.05 \times 10^{-2}$ & $3.74 \times 10^{-2}$/$1.05 \times 10^{-2}$ & $3.67 \times 10^{-2}$/$1.0 \times 10^{-2}$ \\
\cline{1-7}
Bessel & LSTM & False & $7.77 \times 10^{-2}$/$1.96 \times 10^{-2}$ & $2.74 \times 10^{-2}$/$9.08 \times 10^{-3}$ & $2.5 \times 10^{-2}$/$1.21 \times 10^{-2}$ & $1.49 \times 10^{-2}$/$7.69 \times 10^{-3}$ \\
\cline{1-7}
Bessel & LSTM & True & $6.16 \times 10^{-2}$/$1.4 \times 10^{-2}$ & $4.02 \times 10^{-2}$/$1.55 \times 10^{-2}$ & $3.89 \times 10^{-2}$/$1.58 \times 10^{-2}$ & $4.18 \times 10^{-2}$/$1.28 \times 10^{-2}$ \\
\cline{1-7}
Bessel & RNN & False & $4.27 \times 10^{-2}$/$1.0 \times 10^{-2}$ & $1.24 \times 10^{-2}$/$4.43 \times 10^{-3}$ & $7.49 \times 10^{-3}$/$5.23 \times 10^{-3}$ & $6.49 \times 10^{-3}$/$4.95 \times 10^{-3}$ \\
\cline{1-7}
Bessel & RNN & True & $3.77 \times 10^{-2}$/$8.71 \times 10^{-3}$ & $1.69 \times 10^{-2}$/$4.15 \times 10^{-3}$ & $1.61 \times 10^{-2}$/$4.32 \times 10^{-3}$ & $1.62 \times 10^{-2}$/$4.95 \times 10^{-3}$ \\
\cline{1-7}
Damped SHM & GRU & False & $1.27 \times 10^{-1}$/$2.47 \times 10^{-2}$ & $1.88 \times 10^{-2}$/$4.18 \times 10^{-3}$ & $7.89 \times 10^{-3}$/$3.32 \times 10^{-3}$ & $7.17 \times 10^{-3}$/$6.49 \times 10^{-3}$ \\
\cline{1-7}
Damped SHM & GRU & True & $8.34 \times 10^{-2}$/$1.29 \times 10^{-2}$ & $4.52 \times 10^{-2}$/$1.51 \times 10^{-2}$ & $4.42 \times 10^{-2}$/$1.45 \times 10^{-2}$ & $4.51 \times 10^{-2}$/$1.2 \times 10^{-2}$ \\
\cline{1-7}
Damped SHM & LSTM & False & $1.21 \times 10^{-1}$/$2.19 \times 10^{-2}$ & $2.96 \times 10^{-2}$/$9.09 \times 10^{-3}$ & $2.28 \times 10^{-2}$/$8.77 \times 10^{-3}$ & $1.86 \times 10^{-2}$/$1.41 \times 10^{-2}$ \\
\cline{1-7}
Damped SHM & LSTM & True & $1.18 \times 10^{-1}$/$2.37 \times 10^{-2}$ & $5.92 \times 10^{-2}$/$1.9 \times 10^{-2}$ & $5.96 \times 10^{-2}$/$1.72 \times 10^{-2}$ & $6.01 \times 10^{-2}$/$1.75 \times 10^{-2}$ \\
\cline{1-7}
Damped SHM & RNN & False & $2.45 \times 10^{-2}$/$5.18 \times 10^{-3}$ & $1.38 \times 10^{-2}$/$5.48 \times 10^{-3}$ & $1.17 \times 10^{-2}$/$6.43 \times 10^{-3}$ & $8.56 \times 10^{-3}$/$4.95 \times 10^{-3}$ \\
\cline{1-7}
Damped SHM & RNN & True & $7.24 \times 10^{-2}$/$1.72 \times 10^{-2}$ & $1.91 \times 10^{-2}$/$6.21 \times 10^{-3}$ & $1.76 \times 10^{-2}$/$6.8 \times 10^{-3}$ & $1.94 \times 10^{-2}$/$5.47 \times 10^{-3}$ \\
\cline{1-7}
\hline
\end{tabular}}
\caption{{\bfseries Summary of Simulation Results with Quantum Noise Model}}
\label{tableNoisyMSE}
\end{table}
\subsection{Time-Series Prediction-NARMA benchmark}
We further investigate the time-series prediction task with NARMA benchmarks (described in \sectionautorefname{\ref{sec:time_series_prediction}}).
\subsubsection{QRNN}
For QRNN, we observe that in both the NARMA5 and NARMA10 cases (shown in \figureautorefname{\ref{fig:rnn_NARMA5}} and \figureautorefname{\ref{fig:rnn_NARMA10}}), the QRNN learns more structure of the data in the first training epoch. However, we can see that the QRNN-RC can catch up pretty quickly. After 15 epochs of training, the results from QRNN-RC are very close to QRNN.
If we compare the performance of QRNN-RC to classical RNN-RC and RNN, we can see that QRNN-RC provides results superior than classical models with a similar number of parameters. 
%
% \begin{figure}[hbtp]
% \includegraphics[width=1.\linewidth]{results/RNN/RNN_NARMA2_PERFORMANCE_COMPARISON_GRID_full.pdf}% Here is how to import EPS art
% \caption{Learning the NARMA2 with QRNN-RC.}
% \label{fig:rnn_NARMA2}
% \end{figure}
%
\begin{figure}[hbtp]
\includegraphics[width=1.\linewidth]{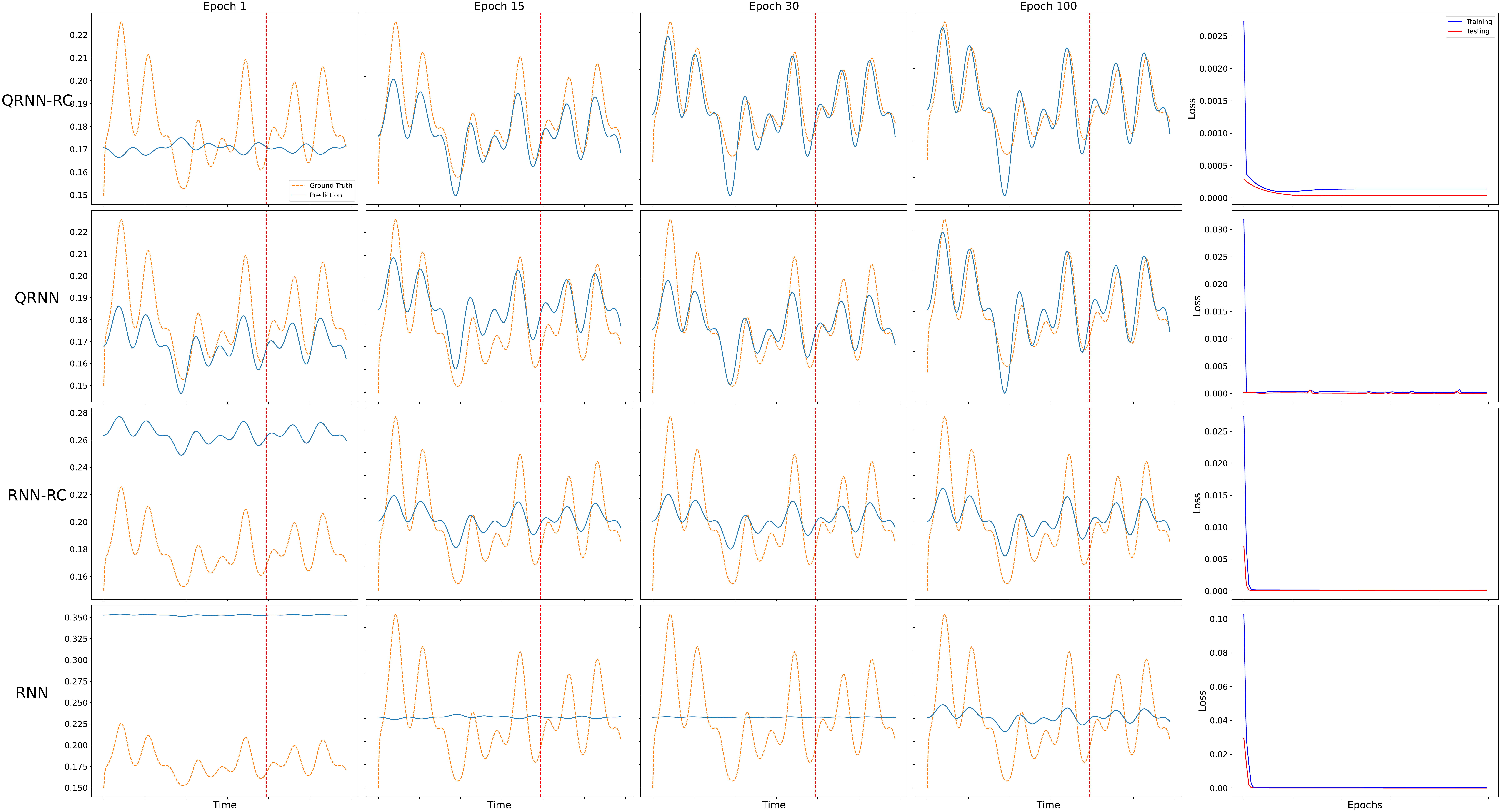}% Here is how to import EPS art
\caption{{\bfseries Learning the NARMA5 with QRNN-RC.} 
%
% We observer that the QRNN learns more structure of the data in the first training epoch. However, we can see that the QRNN-RC can catch up pretty quickly. After 15 epochs of training, the results from QRNN-RC are very close to QRNN and after 30 epochs of training, the QRNN-RC beats QRNN.
%
% If we compare the performance of QRNN-RC to classical RNN-RC and RNN, we can see that QRNN-RC provides results superior than classical models with similar number of parameters. 
%
% The orange dashed line represents the ground truth of the target sequence in NARMA5 while the blue solid line is the output from the models. The vertical red dashed line separates the \emph{training} set (left) from the \emph{testing} set (right).
}
\label{fig:rnn_NARMA5}
\end{figure}
\begin{figure}[hbtp]
\includegraphics[width=1.\linewidth]{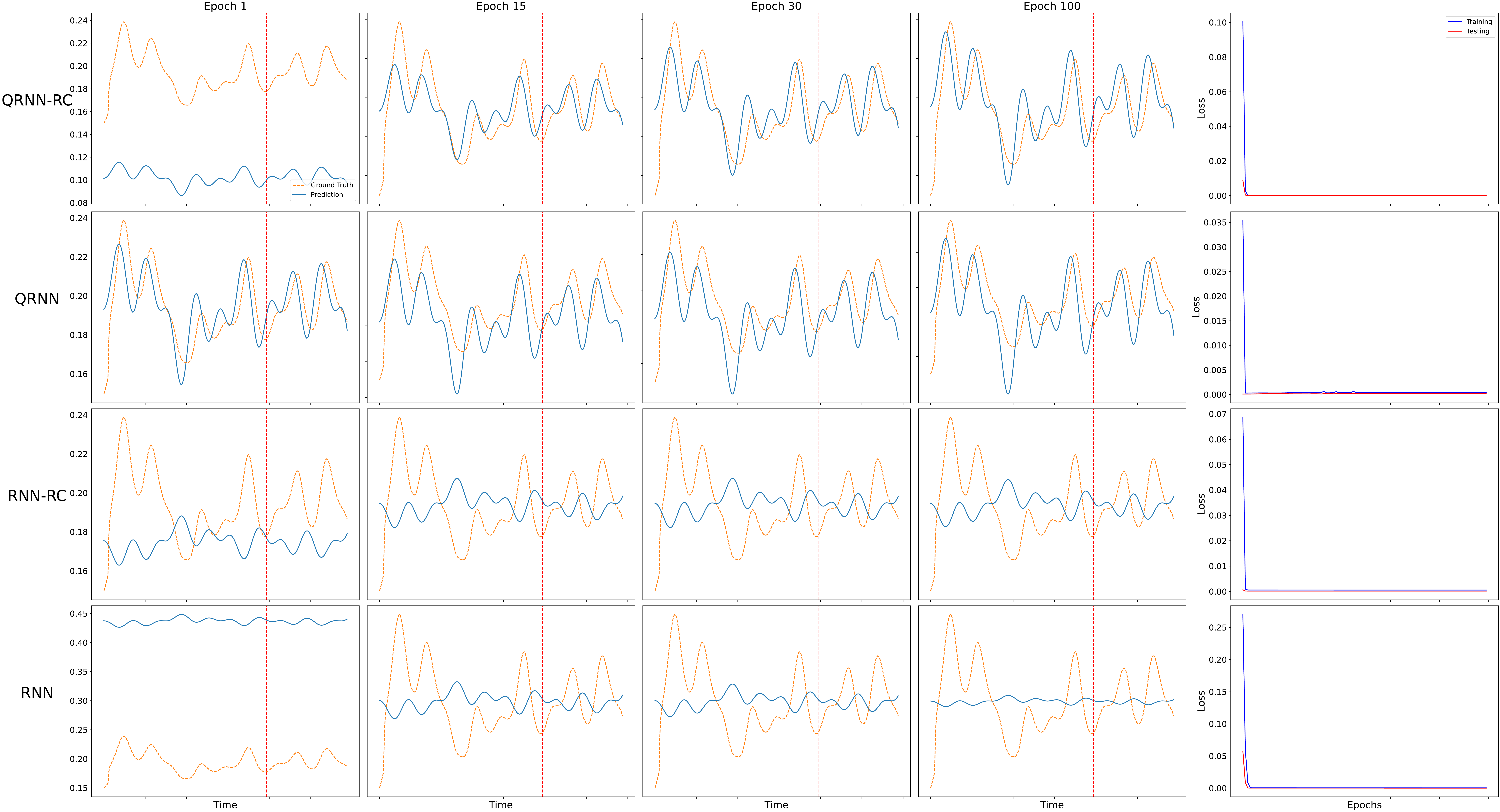}% Here is how to import EPS art
\caption{{\bfseries Learning the NARMA10 with QRNN-RC.}  
% We observer that the QRNN learns more structure of the data in the first training epoch. However, we can see that the QRNN-RC can catch up pretty quickly. After 15 epochs of training, the results from QRNN-RC are very close to QRNN.
%
% If we compare the performance of QRNN-RC to classical RNN-RC and RNN, we can see that QRNN-RC provides results superior than classical models with similar number of parameters. 
%
% The orange dashed line represents the ground truth of the target sequence in NARMA10 while the blue solid line is the output from the models. The vertical red dashed line separates the \emph{training} set (left) from the \emph{testing} set (right).
}
\label{fig:rnn_NARMA10}
\end{figure}
%
% \begin{figure}[hbtp]
% \includegraphics[width=1.\linewidth]{results/GRU/GRU_NARMA2_PERFORMANCE_COMPARISON_GRID_full.pdf}% Here is how to import EPS art
% \caption{Learning the NARMA2 with QGRU-RC.}
% \label{fig:gru_NARMA2}
% \end{figure}
%
\subsubsection{QGRU}
For the QGRU, we observe that in both the NARMA5 and NARMA10 cases (shown in \figureautorefname{\ref{fig:gru_NARMA5}} and \figureautorefname{\ref{fig:gru_NARMA10}}), the QGRU learns more structure of the data in the first training epoch. However, we can see that the QGRU-RC can catch up pretty quickly. After 15 epochs of training, the results from QGRU-RC are very similar to the ones from QGRU. We also see that they are indistinguishable after 100 epochs of training.
In addition, the simulation shows that the performance of QGRU-RC is superior than the classical GRU-RC and GRU with a similar number of parameters.
\begin{figure}[hbtp]
\includegraphics[width=1.\linewidth]{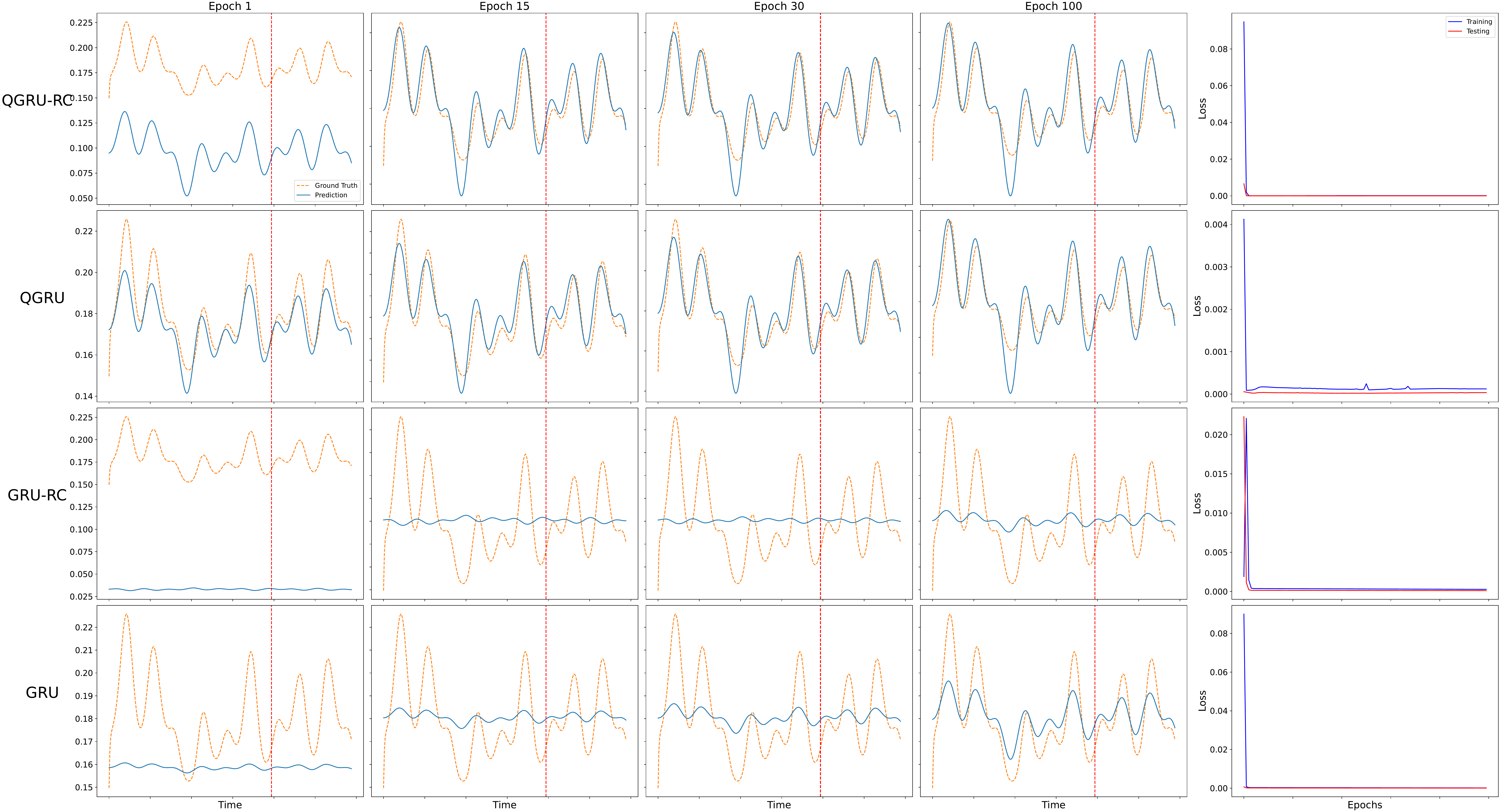}% Here is how to import EPS art
\caption{{\bfseries Learning the NARMA5 with QGRU-RC.} 
%
% We observe that the QGRU learns more structure of the data in the first training epoch. However, we can see that the QGRU-RC can catch up pretty quickly. After 15 epochs of training, the results from QGRU-RC and QGRU are very close.
%
% If we further compare the performance between QGRU-RC and classical models, we can find that QGRU-RC beats classical GRU-RC and GRU with similar number of parameters.
%
% The orange dashed line represents the ground truth of the target sequence in NARMA5 while the blue solid line is the output from the models. The vertical red dashed line separates the \emph{training} set (left) from the \emph{testing} set (right).
}
\label{fig:gru_NARMA5}
\end{figure}
\begin{figure}[hbtp]
\includegraphics[width=1.\linewidth]{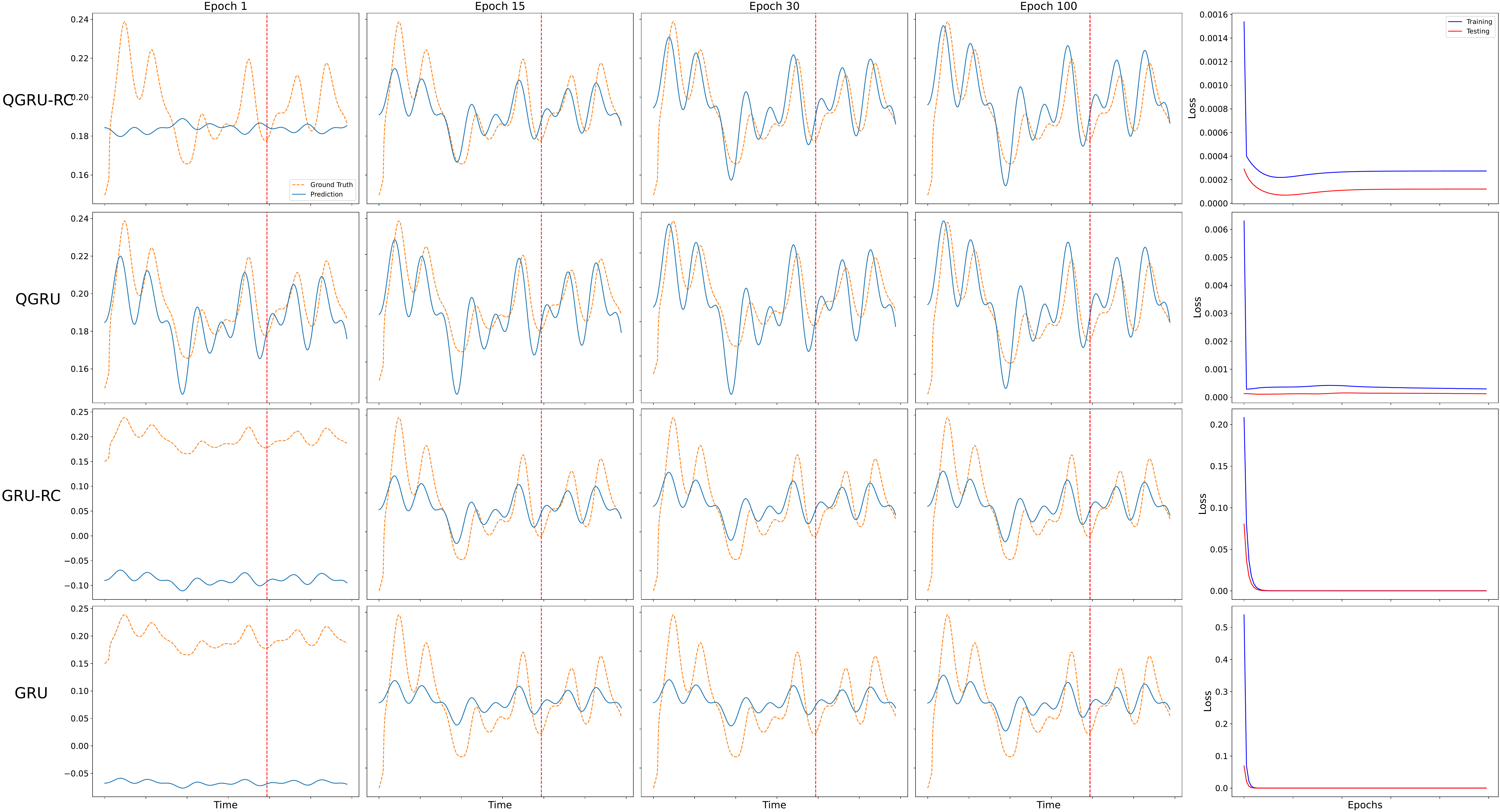}% Here is how to import EPS art
\caption{{\bfseries Learning the NARMA10 with QGRU-RC.} 
%
% We observer that the QGRU learns more structure of the data in the first training epoch. However, we can see that the QGRU-RC can catch up pretty quickly. After 15 epochs of training, the results from QGRU-RC are very similar to the ones from QGRU. We also see that they are indistinguishable after 100 epochs of training.
% %
% In addition, the simulation shows that the performance of QGRU-RC is superior than the classical GRU-RC and GRU with similar number of parameters.
%
% The orange dashed line represents the ground truth of the target sequence in NARMA10 while the blue solid line is the output from the models. The vertical red dashed line separates the \emph{training} set (left) from the \emph{testing} set (right).
}
\label{fig:gru_NARMA10}
\end{figure}
%

% \begin{figure}[hbtp]
% \includegraphics[width=1.\linewidth]{results/LSTM/LSTM_NARMA2_PERFORMANCE_COMPARISON_GRID_full.pdf}% Here is how to import EPS art
% \caption{Learning the NARMA2 with QLSTM-RC.}
% \label{fig:lstm_NARMA2}
% \end{figure}
%
\subsubsection{QLSTM}
For the QLSTM, we observe that in both the NARMA5 and NARMA10 cases (shown in \figureautorefname{\ref{fig:lstm_NARMA5}} and \figureautorefname{\ref{fig:lstm_NARMA10}}), the RC and full optimization of QLSTM can reach good performance after 100 epochs of training. Surprisingly, the training performance of QLSTM-RC is better than the fully optimized one as we can see that the QLSTM-RC predicts the sequence better than QLSTM after 30 epochs of training. 
In addition, we observe that the quantum LSTM, either RC or fully optimized one, perform better than their classical counterparts.
\begin{figure}[hbtp]
\includegraphics[width=1.\linewidth]{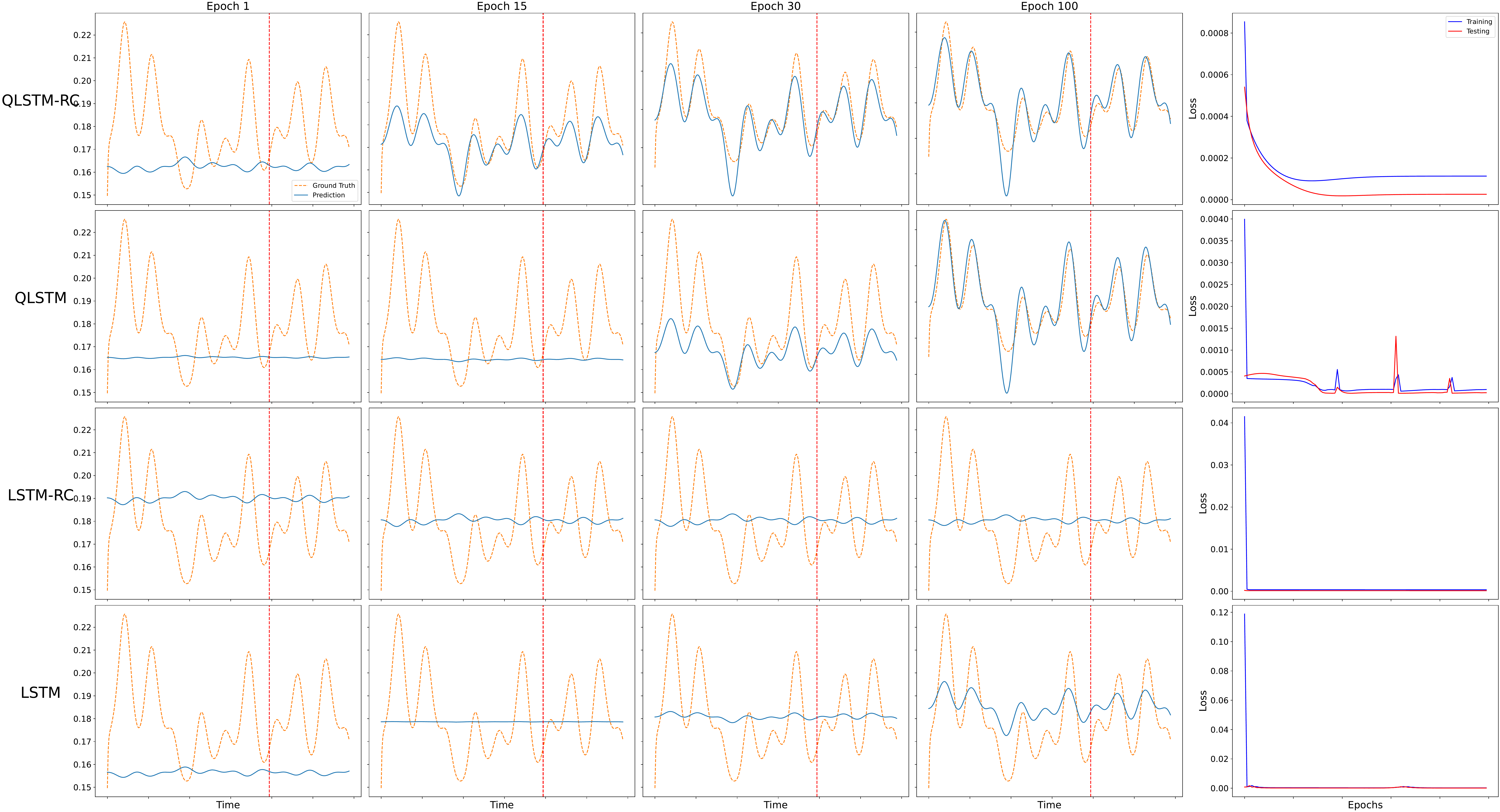}% Here is how to import EPS art
\caption{{\bfseries Learning the NARMA5 with QLSTM-RC.}  
% We observe that both the RC and full optimization of QLSTM can reach good performance after 100 epochs of training. Surprisingly, the training performance of QLSTM-RC is better than the full optimization one as we can see that the QLSTM-RC predicts the sequence better than QLSTM after 30 epochs of training. 
%
% In addition, we observe that the quantum LSTM, either RC or fully optimized one, beat their classical counterparts.
%
% The orange dashed line represents the ground truth of the target sequence in NARMA5 while the blue solid line is the output from the models. The vertical red dashed line separates the \emph{training} set (left) from the \emph{testing} set (right).
}
\label{fig:lstm_NARMA5}
\end{figure}
\begin{figure}[hbtp]
\includegraphics[width=1.\linewidth]{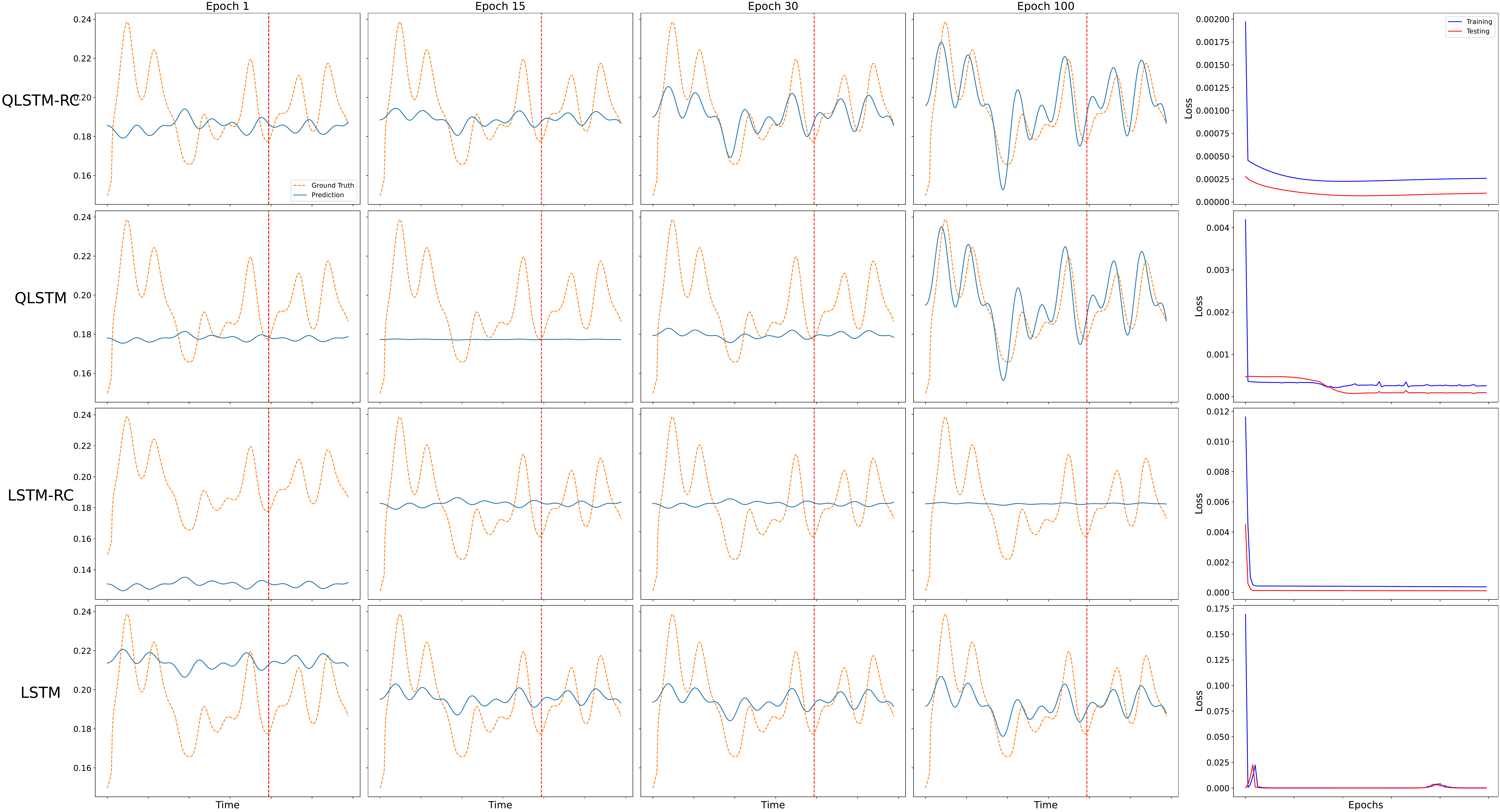}% Here is how to import EPS art
\caption{{\bfseries Learning the NARMA10 with QLSTM-RC.}
%
% We observe that both the RC and full optimization of QLSTM can reach good performance after 100 epochs of training. Surprisingly, the training performance of QLSTM-RC is better than the full optimization one as we can see that the QLSTM-RC predicts the sequence better than QLSTM after 30 epochs of training. 
%
% In addition, we observe that the quantum LSTM, either RC or fully optimized one, beat their classical counterparts.
%
% The orange dashed line represents the ground truth of the target sequence in NARMA10 while the blue solid line is the output from the models. The vertical red dashed line separates the \emph{training} set (left) from the \emph{testing} set (right).
}
\label{fig:lstm_NARMA10}
\end{figure}

%

%

%

%
% \YC{Remark on the y-axis for NARMA results due to varying in instability. not-aligned\\}
%
% \YC{First describe the design of numerical simulation. When you use a open-source package, remember to properly cite the github source or the arxiv paper, nowadays, most open-source packages have a arxiv paper for other researchers to cite.}\\
% In this research, we use the following open-source package to perform the simulation. We use the PennyLane~\cite{bergholm2018pennylane} for the construction of quantum circuits.

\section{\label{sec:Discussion}Discussion}
% \subsection{Physical Reservoir}
% \subsection{Tensor Network Reservoir}
\subsection{Quantum Hardware Efficiency}
Quantum hardware efficiency is a quantum algorithm design consideration in which the demands on quantum computing resources are minimized. This is particularly important in the current noisy intermediate-scale quantum (NISQ) era of quantum computing \cite{preskill2018quantum}. In this paper we consider that hardware efficiency is achieved by running fewer quantum circuits.

The RC framework demonstrated in this work is well-suited for NISQ computers because hardware efficiency is improved significantly over the original three QRNNs. The clear reason for this improvement is that efficient training is limited to the final layer, meaning that a quantum computer would only be used for generating the outputs for the classical linear layer and the quantum parameters are not trained.

As the RC approach is a hardware efficient approach it reduces the negative effects of noise on the quantum computation and therefore can improve the performance of time-series prediction. In our work, the noisy simulation results in Figures~\ref{fig:noisy_rnn}, ~\ref{fig:noisy_gru}, and~\ref{fig:noisy_lstm} show that the RC approach, when compared with the original QRNN algorithm, has smoother prediction curves that are less corrupted by simulation noise. This is highly desirable given that the target function is smooth. In addition there is evidence that the MSE loss curves, particularly for QRNN-RC in \figureautorefname{\ref{fig:noisy_rnn}}, has less noise and stabilizes to a loss minimum in fewer epochs.

\subsection{Potential Applications}

In order to facilitate maximal advantage of a quantum approach to machine learning, the method proposed in this paper can be utilized to decrease the time and complexity required by existing methods for certain applications. In this paper, we analyzed examples of function approximation and time series prediction tasks. This method can further be applied to nuanced tasks using sequential or temporal data, such as using acoustic models for time series classification as implemented in \cite{yang2021voice2series}, facial recognition systems \cite{easom2020towards}, and natural language processing \cite{di2022dawn}. Additionally, there are numerous financial applications \cite{egger2020financial} including time series prediction \cite{krollner2010financial, dingli2017financial} for stock price and market behavior, and classification problems for risk and fraud detection.

% \subsection{Multi-task Learning}
% %
% Multi-task learning (MTL) \cite{ruder2017overview} is a machine learning framework in which representations are shared between related tasks. \YC{MTL definition here.}
% %
% Since the QRNN are not trained and only the last linear layer is trained. The architecture can be used to solve multiple tasks with moderate training resources required. 
% %
% \YC{MTL: using same architecture while optimizing several different loss functions.\\}
% \YC{MTL: training with related tasks\\}
% \YC{MTL: We can see multi-task learning as a form of inductive transfer. The inductive bias introduced in the MTL is that the model will prefer hypotheses which can explain more than one tasks. It can be shown that this usually leads to better generalizability.\\}
% %
% \YC{MTL: hard parameter sharing\\}
% Hard parameter sharing is the most commonly used approach in MTL. The idea is to share the hidden layers between all tasks, while each specific task has its output layers \cite{}. It has been show in the work \cite{} that hard parameter sharing can greatly reduce the risk of overfitting.
% %
% \YC{MTL: soft parameter sharing\\}
%

% \subsection{RC as the Generative Model or Privacy-Preserving Model}

\section{\label{sec:Conclusion}Conclusion}
In this paper, we introduce the function approximation and time-series prediction framework in which the quantum RNN and its variants, such as quantum GRU and quantum LSTM, are used as the reservoir. We show via numerical simulations that the QRNN-RC can reach results comparable to fully trained QRNN models in several function approximation and time-series prediction tasks. Since the QRNNs in the proposed model do not need to be trained, the overall process is much faster than the fully trained ones. We also compare to classical RNN-based RC and show that the quantum solutions require fewer training epochs in most cases. Our results demonstrate a new possibility to utilize quantum neural networks for sequential modeling with very small amount of resource requirement. 

\clearpage
\begin{acknowledgments}
The authors would like to thank Constantin Gonciulea and Vanio Markov for constructive and helpful discussions during the development of this paper.
The views expressed in this article are those of the authors and do not represent the views of Wells Fargo. This
article is for informational purposes only. Nothing contained in this article should be construed as investment advice.
Wells Fargo makes no express or implied warranties and expressly disclaims all legal, tax, and accounting implications
related to this article.\\
\end{acknowledgments}

\appendix

% \section{Hyperparameters}

\bibliographystyle{ieeetr}
\bibliography{apssamp,bib/qml_examples,bib/reservoir_computing,bib/qc_basic,bib/time_series_ref,bib/classical_ml_cv,bib/classical_ml_nlp,bib/rnn_classic,bib/vqc,bib/classical_ml_rl,bib/qml_basic_review,bib/tool,bib/classical_ml_finance,bib/multi_task_learning,bib/noise_simulation,bib/classical_ml,bib/narma_related}% Produces the bibliography via BibTeX.

\end{document}